%% file: _main.tex
\documentclass[letterpaper,10pt,conference]{formatting/ieeeconf}
\IEEEoverridecommandlockouts
\let\labelindent\relax
\usepackage{amsfonts}       

\usepackage{amsthm}
\usepackage{mathtools}      
\usepackage{amssymb}        
\usepackage{filecontents}
\usepackage{graphicx}       
\usepackage{subcaption}
\usepackage{marginnote}     
\usepackage{marvosym}       
\usepackage{overpic}        
\usepackage{tabularx}
\usepackage{booktabs}
\usepackage{cite}
\usepackage{color}
\usepackage[linesnumbered,algoruled,boxed,lined]{algorithm2e}
\usepackage[normalem]{ulem}
\usepackage{epstopdf}
\usepackage{enumitem}
\PassOptionsToPackage{hyphens}{url}
\usepackage{xurl}
\usepackage[a-2b,mathxmp]{pdfx}[2018/12/22]
\hypersetup{hidelinks}

\newif\ifdraft
\draftfalse

\ifdraft
\usepackage[paperheight=11in,paperwidth=9.5in,
			left=1.25in,right=1.25in,
			top=0.75in,bottom=0.75in,
			heightrounded,marginparwidth=1.2in,
			marginparsep=0.05in]{geometry}
\usepackage{xcolor}
\usepackage{xargs} 
\usepackage[textsize=footnotesize]{todonotes}
\newcommandx{\nt}[2][1=]{\todo[linecolor=red,
			backgroundcolor=red!10,bordercolor=red,#1]{#2}}
\newcommandx{\jy}[2][1=]{\todo[linecolor=green,
			backgroundcolor=green!10,bordercolor=green,#1]{JY:#2}}
\else
\newcommand{\nt}[1]{{}}
\newcommand{\jy}[1]{{}}
\fi

\newif\iftwocolumn
\twocolumntrue

\theoremstyle{definition}

\theoremstyle{remark}

\SetKwProg{Fn}{Function}{}{}
\SetKwComment{Comment}{$\triangleright$\ }{}

\makeatletter
\def\section{\@startsection{section}{1}{\z@}{0.75ex plus 0.5ex minus 0.25ex}
  {0.35ex plus 0.25ex minus 0.1ex}{\normalfont\normalsize\centering\scshape}}
\def\subsection{\@startsection{subsection}{2}{\z@}{0.75ex plus 0.5ex minus 0.25ex}
  {0.35ex plus 0.25ex minus 0.1ex}{\normalfont\normalsize\itshape}}
\def\subsubsection{\@startsection{subsubsection}
                                 {3}
                                 {\z@ \hspace*{1mm}}
                                 {0ex plus 0.1ex minus 0.1ex}
                                 {0ex}
                                 {\normalfont\normalsize\itshape}}
\makeatother

\input{texs/benchmark_numbers}
\input{texs/momentum_numbers}

\input{texs/computational_numbers}

\title{\fontsize{16.5}{24}\selectfont Efficient B\'ezier Velocity Optimization for Free-Floating Space Manipulators}
\author{Duo Zhang \qquad Zhizhuo Zhang \qquad Xiaoli Bai \qquad Jingjin Yu%
\thanks{D. Zhang and J. Yu are with the Department of Computer Science, and
Z. Zhang and X. Bai are with the Department of Mechanical and Aerospace Engineering,
Rutgers, the State University of New Jersey, Piscataway, NJ, USA.
E-Mails: \texttt{\{duo.zhang, zhizhuo.zhang, xiaoli.bai, jingjin.yu\}@rutgers.edu}.
This material is based upon work supported by the Air Force Office of Scientific
Research under award number FA9550-25-1-0249.}}

\begin{document}
\flushbottom

\maketitle
\thispagestyle{empty}
\pagestyle{empty}


\begin{abstract}
We present FAVOR (Free-floating Arm Velocity Optimization with Recursive
Sensitivities), a planner for collision-free reaching, tracking, and
prescribed-time pre-grasp interception on an unactuated spacecraft.
It optimizes B\'ezier joint-velocity curves with linear velocity,
acceleration, and continuity constraints. Decision dimension is independent
of rollout resolution. Analytical recursive sensitivities provide task and
clearance gradients through the coupled base--arm motion. Parallel
evaluation, caching, and incremental collision discovery reduce computation.
With a seven-DoF arm and five simulated spacecraft models, FAVOR achieves
\PointIpoptRate\% point-to-point success with
\PointIpoptTime~s mean computation, versus \PointSplineRate\% and
\PointSplineTime~s for an IK-initialized position-spline baseline. Means
include failures and timeouts. FAVOR completes 30 of 36 tracking cases,
versus 15 for single-step QP, and all \InterceptionTotal{} interception
instances, versus \InterceptionSplineSuccess{} for the spline baseline.
A controlled ablation shows that finite differences increase mean planning
time \CompFiniteDifferenceSlowdown{}-fold.
\end{abstract}

\section{Introduction}\label{sec:intro}
\input{texs/00-intro}

\section{Related Work}\label{sec:related}
\input{texs/01-related}

\section{Model and Planning Formulation}\label{sec:problem}
\input{texs/03-prelim}
\input{texs/04-problem}

\section{Trajectory Optimization}\label{sec:analysis}
\input{texs/06-analysis}
\input{texs/09-algorithm}

\section{Evaluation}\label{sec:evaluation}
\input{texs/12-evaluation}

\section{Conclusion}\label{sec:conclusion}
\input{texs/15-conclusion}

{\small
\bibliographystyle{formatting/IEEEtran}
\bibliography{bib/jingjin}
}

\end{document}

%% file: texs/benchmark_numbers.tex
\newcommand{\PointTotal}{625}
\newcommand{\PointIkSuccess}{601}
\newcommand{\PointIkTime}{4.130}
\newcommand{\PointInitializationShared}{600}
\newcommand{\PointIkSharedTime}{1.919}
\newcommand{\PointJacobianSharedTime}{1.661}
\newcommand{\TrackingBudget}{4}

\newcommand{\TrackingSeedSuccess}{19}
\newcommand{\PointIpoptBudget}{60}

\newcommand{\PointIpoptSuccess}{624}
\newcommand{\PointIpoptRate}{99.8}
\newcommand{\PointIpoptTime}{1.779}

\newcommand{\PointSplineRate}{62.9}
\newcommand{\PointSplineTime}{50.270}

\newcommand{\PointPositionMeanMM}{1.90}
\newcommand{\PointOrientationMean}{0.0094}

\newcommand{\InterceptionTotal}{36}
\newcommand{\InterceptionCandidateSuccess}{143}
\newcommand{\InterceptionTimeMean}{3.043}
\newcommand{\InterceptionPositionMeanMM}{1.13}
\newcommand{\InterceptionOrientationMean}{0.0025}
\newcommand{\InterceptionSplineSuccess}{22}
\newcommand{\InterceptionSplineTime}{34.824}

\newcommand{\TrackingStaticIpoptSuccess}{23}

\newcommand{\TrackingStaticLegacyIkSuccess}{23}

\newcommand{\TrackingStaticQpSuccess}{10}

\newcommand{\TrackingStaticShared}{10}
\newcommand{\TrackingDynamicIpoptSuccess}{7}

\newcommand{\TrackingDynamicLegacyIkSuccess}{6}

\newcommand{\TrackingDynamicQpSuccess}{5}

\newcommand{\TrackingDynamicShared}{5}

%% file: texs/momentum_numbers.tex
\newcommand{\MomentumTrajectories}{877}
\newcommand{\MomentumPMaximum}{\ensuremath{3.85\times10^{-14}}}

\newcommand{\MomentumHMaximum}{\ensuremath{1.26\times10^{-13}}}

%% file: texs/computational_numbers.tex
\newcommand{\CompCases}{125}
\newcommand{\CompFullSuccess}{125}

\newcommand{\CompFiniteDifferenceSuccess}{115}

\newcommand{\CompFiniteDifferenceSlowdown}{4.7}
\newcommand{\CompJacobianSpeedup}{6.9}

%% file: texs/00-intro.tex
Every reach of a free-floating space manipulator is also a maneuver of the
spacecraft that carries it. Moving the arm redistributes momentum and can
translate and rotate the unactuated base, changing both the end-effector
pose and the clearance to surrounding structures. We seek collision-free
joint-velocity commands that bring the end effector to a desired pose at
a specified time, given the initial state and momentum, while respecting
task tolerances and joint limits. Planning with these base reactions can
reduce the need for propellant-consuming corrections
\cite{bhundiya2026propellant} and support manipulation when base actuation
is unavailable.

Robotic access to spacecraft already in orbit offers a route to longer
missions and the removal of debris. Astroscale's ADRAS-J flew around an
existing rocket upper stage, imaging its motion and structural condition
in preparation for future removal \cite{astroscale2024adras}. On the
International Space Station, NASA's Robotic Refueling Mission used Dextre
to cut wires, remove protective caps, and demonstrate fluid transfer through
a satellite-like servicing interface \cite{nasa2013rrm}. ESA's planned
ClearSpace-1 mission aims to capture PROBA-1, an unprepared spacecraft
without dedicated capture interfaces \cite{metrailler2026clearspace}.
Bringing a tool to a servicing interface or a gripper to a moving object
requires precise control of the approach: surrounding structures constrain
the route, and target motion constrains the timing.

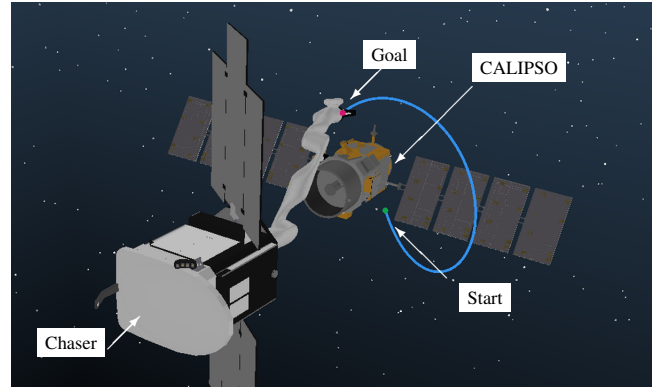
\begin{figure}[t]
  \centering
  \sbox0{\includegraphics{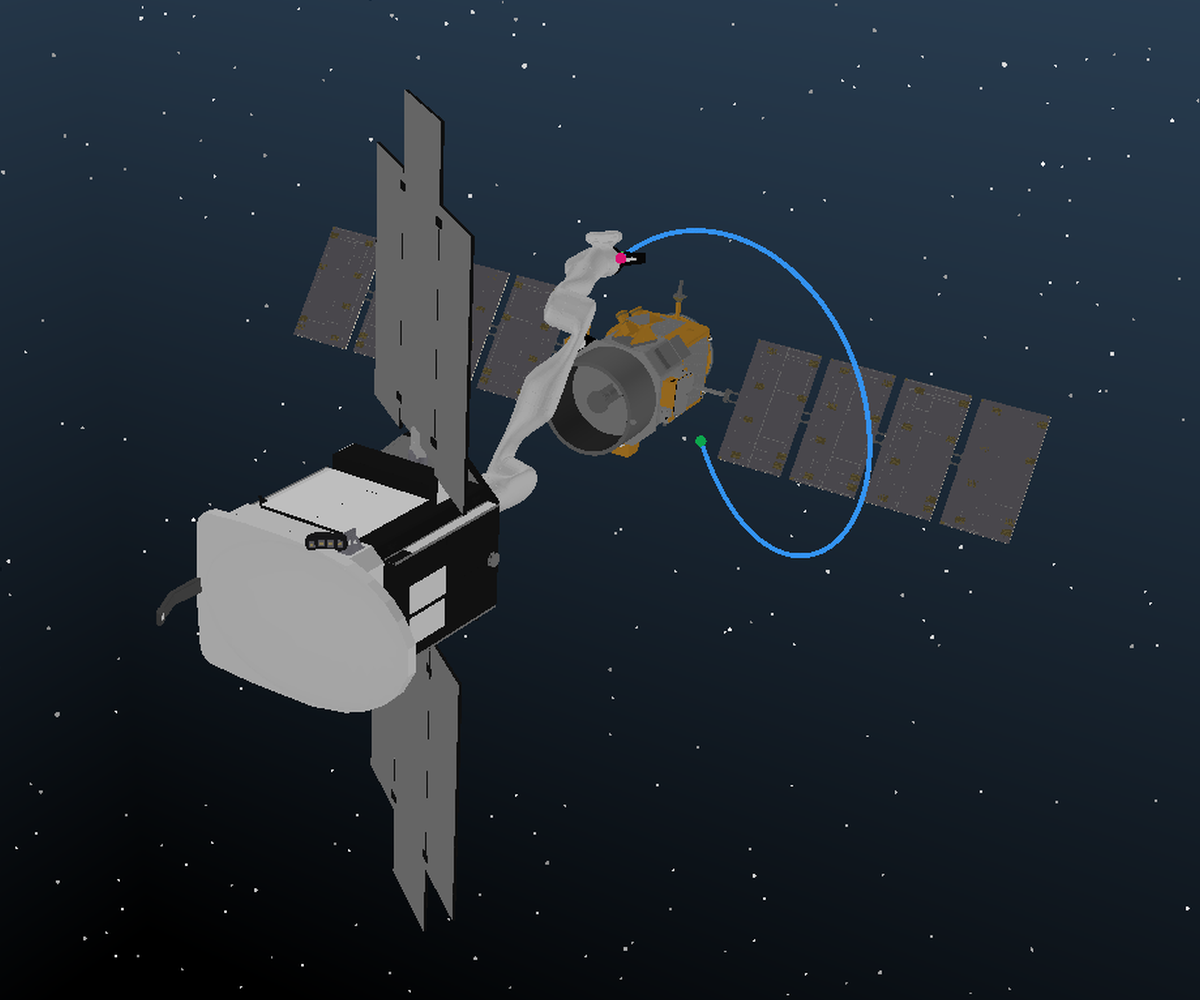}}
  \edef\figuretoptrim{\the\dimexpr\ht0/20\relax}
  \edef\figurebottomtrim{\the\dimexpr\ht0*23/100\relax}
  \begin{overpic}[width=0.98\columnwidth,
    trim=0pt \figurebottomtrim{} 0pt \figuretoptrim{},clip]{figures/trajectory_point_to_point.png}
    \put(55,50.8333){\colorbox{white}{\scriptsize Goal}}
    \put(58,49.8333){\color{white}\vector(-1,-1){5}}
    \put(70,12.8333){\colorbox{white}{\scriptsize Start}}
    \put(68,16.8333){\color{white}\vector(-1,1){8}}
    \put(72,48.8333){\colorbox{white}{\scriptsize CALIPSO}}
    \put(72,46.8333){\color{white}\vector(-1,-1){12}}
    \put(4,5.8333){\colorbox{white}{\scriptsize Chaser}}
    \put(15,5.8333){\color{white}\vector(1,1){5}}
  \end{overpic}
  \caption{A planned detour around CALIPSO. Given the initial free-floating
  state, goal pose, and arrival time, the planner determines the blue
  end-effector path and coupled base--arm motion. Green and magenta mark
  start and arrival; the robot is shown at arrival.}
  \vspace{-4mm}
  \label{fig:task-overlays}
\end{figure}

Our CALIPSO inspection tasks bring the end effector to the Imaging
Infrared Radiometer (IIR), Integrated Lidar Receiver (ILR), and Star Tracker
Assembly (STA) \cite{nasa2005essp,weimer2015qualification}. Solar arrays and
telescope geometry can obstruct direct approaches to these regions.
The detour in Fig.~\ref{fig:task-overlays} illustrates how reaching an
instrument requires coordinating the end-effector path with the motion
of the entire chaser base--arm system.

The consequences of a joint command persist throughout the maneuver:
momentum conservation couples it to later base poses, task errors, and
collision clearances. Nonzero initial momentum adds drift even when the
joints stop. Generalized-Jacobian and trajectory-optimization methods
account for this coupling
\cite{misra2017task,lampariello2013grasping,rybus2026cartesian}. With detailed
geometry and fine temporal sampling, however, repeatedly simulating
candidate motions and differentiating their constraints can dominate the
planning cost. The computational challenge is to retain the full coupled
motion while making its optimization tractable.

We introduce \textbf{FAVOR}---\emph{Free-floating Arm Velocity Optimization
with Recursive Sensitivities}---to address this challenge. FAVOR optimizes
piecewise B\'ezier joint-velocity curves, with the momentum-coupled rollout
determining the Cartesian trajectory and base motion. Linear coefficient
constraints enforce velocity, acceleration, and segment continuity
requirements; refining the rollout adds no decision variables. Analytical
recursive sensitivities propagate control effects through the trajectory
to compute task and clearance gradients. Parallel local derivatives and
shared caching reduce repeated computation, while incremental collision
discovery adds constraints for nearby geometry. The formulation supports
endpoint planning, reference tracking, and pre-grasp interception with
prescribed object motion and arrival time.

In summary, our key technical contributions are: 
\begin{itemize}[leftmargin=*,nosep]
  \item We propose a unified approach to prescribed-time free-floating
  manipulation across three task classes, using compact joint-velocity
  optimization with linear velocity, acceleration, and continuity constraints.
  \item We develop analytical task and collision derivatives through the
  coupled rollout, combining recursive sensitivities, parallel local
  derivatives, and shared caching for efficient optimization of coupled
  base--arm trajectories.
  \item We evaluate FAVOR across spacecraft geometries, initial motions,
  and base-to-arm mass ratios, and isolate the computational benefits of
  analytical derivatives, caching, and recursive assembly through controlled
  ablation studies.
\end{itemize}

Across simulations with a seven-DoF arm, FAVOR solves
\PointIpoptSuccess/\PointTotal{} point-to-point cases
(\PointIpoptRate\%) with \PointIpoptTime~s mean computation, compared with
\PointSplineRate\% and \PointSplineTime~s for the adapted position-spline
baseline initialized by inverse kinematics. These means include failed
attempts and timeouts. FAVOR completes 30 of 36 tracking cases, twice the
15 completed by the single-step generalized-Jacobian QP. It also solves all
\InterceptionTotal{} pre-grasp interception instances, compared with
\InterceptionSplineSuccess{} for the IK-initialized spline baseline.
Mean interception computation is \InterceptionTimeMean~s versus
\InterceptionSplineTime~s. A controlled \CompCases{}-case ablation isolates
the impact of analytical derivatives:
replacing analytical derivatives with finite differences increases mean
planning time \CompFiniteDifferenceSlowdown{}-fold and reduces successes
from \CompFullSuccess{} to \CompFiniteDifferenceSuccess{}.

%% file: texs/01-related.tex
\paragraph{Free-Flying Planning with Base Actuation}

Convex optimization supports endpoint planning with base attitude control
and capture--detumble maneuvers with coordinated base--arm actuation
\cite{misra2017optimal,virgili2019capture}. Learned warm-starts accelerate
sequential convex terminal-approach planning with commanded spacecraft
forces, spacecraft torques, and joint torques \cite{takubo2026transformer}.
These formulations exploit base actuation; our free-floating setting uses
joint-velocity commands with conserved system momentum.

\paragraph{Free-Floating Motion Planning}

Sampling-based planners explore alternative routes through A*, RRT, and
RRT* variants \cite{persson2014sampling,rybus2015rrt,rybus2020rrt,liu2025rrtstar}.
Steering and propagation must account for base--arm coupling. Learning-based
and hybrid methods provide reaching policies and combine geometric search
with joint control for fixed or moving targets
\cite{wu2020rl,wang2022hdo}.
Trajectory optimization searches over joint-position splines, B\'ezier curves,
or joint accelerations under task and collision constraints
\cite{lampariello2013grasping,rybus2022spline,wang2018pso,wang2018de,li2023pseudospectral}.
For general articulated robots, composite B\'ezier parameterizations combine
with conservative motion bounds and adaptive subdivision to maintain
collision feasibility \cite{zhang2024provably}.
Base--arm coupling also enables manipulator-driven spacecraft attitude
maneuvers \cite{bhundiya2026propellant}. Generalized-Jacobian and successive-QP
formulations recover joint motion from optimized Cartesian curves or
prescribed tracking and approach paths
\cite{rybus2026cartesian,misra2017task,li2026softcapture}.
FAVOR directly optimizes bounded joint-velocity curves and
propagates the coupled system to determine the Cartesian route.
Collision-aware trajectory optimization uses swept convex checks
\cite{schulman2014motion} and semi-infinite formulations to generate incremental
constraints \cite{hauser2021semiinfinite}.

%% file: texs/03-prelim.tex
\subsection{Momentum-Coupled Motion}

Consider an $n$-DoF arm mounted on an unactuated spacecraft. No external wrench
acts during the planned maneuver, so the total linear and angular momenta,
$P_0,H_0\in\mathbb{R}^3$, are conserved. Angular momentum $H_0$ is
measured about the fixed inertial origin. Let $r_b$ and $R_b\in SO(3)$ be the
base position and attitude, and let $\theta\in\mathbb{R}^n$ denote the joints.
Base linear and angular velocities are expressed in the inertial frame.
With the joint velocity
$u=\dot\theta$ as input, the momentum relation can be partitioned as
\begin{equation}
  M_b(q)\begin{bmatrix}v_b\\\omega_b\end{bmatrix}
  +M_a(q)u
  =\begin{bmatrix}P_0\\H_0-r_b\times P_0\end{bmatrix},
  \label{eq:momentum-partition}
\end{equation}
where $M_b\in\mathbb{R}^{6\times6}$ and $M_a\in\mathbb{R}^{6\times n}$
map base and joint velocities, respectively, to system momentum. Both
contain the base and link mass, inertia, and kinematic contributions
\cite{misra2017task}. The term $r_b\times P_0$ shifts angular momentum
from the base reference point to the inertial origin. Solving for the
base velocity yields
\begin{equation}
 \begin{bmatrix}v_b\\\omega_b\end{bmatrix}
 =\underbrace{-M_b^{-1}M_a}_{B(q)}u+
 \underbrace{M_b^{-1}\begin{bmatrix}P_0\\H_0-r_b\times P_0\end{bmatrix}}_{b(q)}.
 \label{eq:base-affine-map}
\end{equation}
Equation~\eqref{eq:base-affine-map} determines the base reaction from the
joint command and current configuration. Its coefficients change along the
motion, making the full trajectory nonlinear in the control history.
The drift term $b(q)$ also permits base motion at zero joint velocity when
the initial momentum is nonzero.

We represent the configuration as
$q=[r_b^T,\eta_b^T,\theta^T]^T\in\mathbb{R}^{6+n}$, where $\eta_b$ denotes
local Euler-XYZ attitude coordinates; attitude propagation uses rotation matrices.
The end-effector pose $(r_e,R_e)$ is obtained by forward kinematics of the
complete base-and-arm configuration.

\subsection{Compact Joint-Velocity Representation}

FAVOR keeps the decision dimension independent of rollout resolution by
parameterizing each joint-velocity history with $S$ B\'ezier segments.
Segment $s$ has duration $T_s=T/S$ and degree $d$:
\begin{equation}
 u_s(\sigma)=\sum_{i=0}^{d}\binom{d}{i}(1-\sigma)^{d-i}\sigma^iP_{s,i},
 \qquad \sigma\in[0,1].
 \label{eq:bezier-control}
\end{equation}
Here $\sigma=(t-(s-1)T_s)/T_s$ is normalized time in segment
$s=1,\ldots,S$, and $P_{s,i}\in\mathbb{R}^n$ are joint-velocity control points.
The convex-hull
property gives continuous-time velocity bounds by bounding the control points.
Similarly, the derivative is a degree-$(d-1)$ B\'ezier curve with control points
\begin{equation}
 D_{s,i}=\frac{d}{T_s}(P_{s,i+1}-P_{s,i}),
 \label{eq:bezier-derivative}
\end{equation}
so bounds on $D_{s,i}$ constrain acceleration within the planned input
\cite{farin2002curves}.

Adjacent segments share their boundary velocity, $P_{s,d}=P_{s+1,0}$. Matching first
derivatives at a join additionally gives $C^1$ velocity, or continuous
acceleration; matching second derivatives gives $C^2$ velocity, or
continuous jerk. These endpoint conditions are linear in the control
points. Joint positions follow by integration.
For a chosen velocity continuity order $m$, the join conditions are built
into a sparse linear map from independent parameters $z$,
\begin{equation}
 \operatorname{vec}(\mathcal P)=A_mz.
 \label{eq:control-parameter-map}
\end{equation}
Here $\mathcal P$ collects all segment control points $P_{s,i}$.
The map enforces continuity by construction. For seven-DoF quintic segments
at $C^0$ continuity, $z$ contains $7(5S+1)$ entries. The initial-velocity
boundary fixes seven entries, leaving $35S$ free parameters: 35 for one
segment, 70 for two, and 140 for four. Increasing the number of rollout
samples improves the resolution of state and collision evaluation without
adding control variables.

%% file: texs/04-problem.tex
\subsection{Tasks and Constraints}

The inputs are an initial configuration $q_0$, initial joint velocity
$u_{\mathrm{initial}}\in\mathbb R^n$, conserved momenta $(P_0,H_0)$,
a terminal end-effector pose, and an arrival time $T$.
FAVOR chooses the velocity parameters $z$ for prescribed $T$ and
$\Delta t=T/N$, with sampled commands $u_k=u(k\Delta t)$ and rollout
$q_{k+1}=F(q_k,u_k,\Delta t)$. Here $F$ is the momentum-coupled,
one-step state-update map specified by \eqref{eq:nonlinear-rollout}.
Optimization and reported replay use the same time grid and command
intervals; changing the command application period changes the input schedule.
This leaves the intermediate Cartesian path and terminal joint configuration
free. Tracking adds desired poses along the horizon. Interception supplies
an object-relative pre-grasp pose evaluated at $T$, together with the object's
geometry throughout its motion.

At sample $k$, define the task errors
\begin{equation}
 e_{p,k}=r_{e,k}-r_{d,k},\qquad
 e_{R,k}=\operatorname{Log}(R_{d,k}R_{e,k}^{T}).
 \label{eq:pose-errors}
\end{equation}
The reduced nonlinear program is
\begin{subequations}\label{eq:planner-nlp}
\begin{align}
 \underset{z}{\min}\quad
 &\Phi_T+\Phi_{\mathrm{track}}+\Phi_{\mathrm{ctrl}}+\Phi_{\mathrm{base}},\\
 \text{s.t.}\quad
 &|e_{p,N}^{(j)}|\leq\epsilon_p/\sqrt{3},\quad |e_{R,N}^{(j)}|\leq\epsilon_R/\sqrt{3},\\
 &P_{1,0}=u_{\mathrm{initial}},\quad \theta_{\min}\leq\theta_k\leq\theta_{\max},\\
 &u_{\min}\leq P_{s,i}\leq u_{\max},\quad a_{\min}\leq D_{s,i}\leq a_{\max},\\
 &\delta_{a,k}(q_k,t_k)\geq m_{\mathrm{col}},\quad(a,k)\in\mathcal A.
 \label{eq:nlp-collision}
\end{align}
\end{subequations}
Here $\operatorname{Log}(R)=(\log R)^\vee\in\mathbb R^3$ is the
rotation-vector logarithm. The vee operator is the inverse of the skew-matrix
map: $([v]_\times)^\vee=v$, where $[v]_\times w=v\times w$ for
$v,w\in\mathbb R^3$. For endpoint planning, only the terminal desired pose
is required.
The bounds apply for $j=1,2,3$, $k=0,\ldots,N$, and all segment
coefficients. Here $\theta_{\min}$ and $\theta_{\max}$ are the joint-position limits;
$P_{1,0}=u_{\mathrm{initial}}$ matches the prescribed initial joint velocity.
The factor $1/\sqrt{3}$ ensures each three-component error meets its
Euclidean tolerance, using $\|e\|_2\leq\sqrt{3}\|e\|_\infty$.
The horizon objective uses the sampled costs:
\begin{equation}
 \setlength{\jot}{1pt}
 \begin{aligned}
 \Phi_T&=w_p^T\|e_{p,N}\|^2+w_R^T\|e_{R,N}\|^2,\\
 \Phi_{\mathrm{track}}&=\Delta t\sum\nolimits_{k=0}^{N}
     \left(w_p^r\|e_{p,k}\|^2+w_R^r\|e_{R,k}\|^2\right),\\
 \Phi_{\mathrm{ctrl}}&=\sum\nolimits_{k=0}^{N}
     \left(w_u\|u_k\|^2+w_a\|\dot u_k\|^2\right),\\
 \Phi_{\mathrm{base}}&=w_b\Delta t\sum\nolimits_{k=0}^{N}
     \alpha_k\|\omega_{b,k}\|^2.
 \end{aligned}
 \label{eq:objective-components}
\end{equation}
The trapezoidal factors are $\alpha_0=\alpha_N=1/2$ and $\alpha_k=1$
otherwise. The control term $\Phi_{\mathrm{ctrl}}$ is a sampled regularizer
with weights defined for the fixed grid; the tracking and base-motion terms
approximate time integrals. The nonnegative weights balance task accuracy,
control effort, acceleration, and base angular motion.

\subsection{Collision Representation}

Robot and environment meshes are fully enclosed by conservative sphere unions
constructed using FOAM \cite{coumar2025foam}. Enforcing clearance between these
unions therefore preserves the same clearance for the enclosed meshes.
For a robot sphere with center $c_i(q_k)$ and radius
$\rho_i$, an obstacle sphere has known center $c_o(t_k)$ and radius $\rho_o$.
The signed separations for environment and self-collision are
\begin{align}
 \delta_{io,k}&=\|c_i(q_k)-c_o(t_k)\|-\rho_i-\rho_o,\nonumber\\
 \delta_{ij,k}&=\|c_i(q_k)-c_j(q_k)\|-\rho_i-\rho_j.
 \label{eq:sphere-signed-distances}
\end{align}
Negative values indicate penetration. Self-collision pairs exclude geometries
on the same or adjacent parent joints and internal gripper pairs.
Bounding-volume hierarchies (BVHs) identify nearby pairs, and the sphere distances
provide the narrow-phase constraints. The active set $\mathcal A$ contains
pair--sample combinations discovered during optimization.

For a moving rigid obstacle, its local sphere centers are transformed by
the prescribed pose at each collision-checking time,
$c_o(t_k)=r_o(t_k)+R_o(t_k)c_o^{\mathrm{local}}$.
An environment BVH is cached for each queried time sample and reused across
control evaluations; a static environment shares its BVH across the horizon.
Robot sphere placements and their BVHs are updated with the candidate
trajectory, and each robot BVH is queried against the environment BVH at
the same time. Because arrival time is fixed, obstacle placements are
independent of the velocity parameters, while the robot motion and its
clearance derivatives change during optimization.

%% file: texs/06-analysis.tex
\subsection{Nonlinear Rollout and Analytical Derivatives}

The robot model and conserved momenta $(P_0,H_0)$ are fixed.
The changing momentum map couples each state to preceding joint commands.
Writing the components of \eqref{eq:base-affine-map} as $(v_{b,k},\omega_{b,k})$,
the simulator propagates the configuration by
\begin{align}
 r_{b,k+1}&=r_{b,k}+\Delta t\,v_{b,k},\nonumber\\
 R_{b,k+1}&=\operatorname{Exp}([\Delta t\,\omega_{b,k}]_\times)R_{b,k},\nonumber\\
 \theta_{k+1}&=\theta_k+\Delta t\,u_k.
 \label{eq:nonlinear-rollout}
\end{align}
Together, these updates define the discrete state-transition map
$q_{k+1}=F(q_k,u_k,\Delta t)$, with the updated rotation expressed in the
attitude coordinates $\eta_{b,k+1}$.

At the nominal sample $(q_k,u_k,\Delta t)$, define
$A_k=\partial F/\partial q_k$ and $C_k=\partial F/\partial u_k$ with
$\Delta t$ fixed. The control Jacobian $U_k=\partial u_k/\partial z$
follows directly from the Bernstein basis and shared-boundary map. For the base reaction, write
$\nu=[v_b^T,\omega_b^T]^T$ and let $h$ be the right-hand side of
\eqref{eq:momentum-partition}. Its differential is
$\delta h=[0^T,(-\delta r_b\times P_0)^T]^T$, giving
\begin{equation}
 \delta\nu=M_b^{-1}
   \left(\delta h-\delta M_b\,\nu-\delta M_a\,u-M_a\delta u\right).
 \label{eq:base-velocity-differential}
\end{equation}
Thus $\partial\nu/\partial u=-M_b^{-1}M_a$; varying each coordinate
with $u$ fixed gives $\partial\nu/\partial q$. Link poses, joint axes,
and world-frame inertias provide the analytical derivatives of $M_b$ and $M_a$.

The translational and joint rows follow directly from
\eqref{eq:nonlinear-rollout}. For attitude, write
$Q_k=\operatorname{Exp}([\Delta t\,\omega_{b,k}]_\times)$ and define
$R'_b=\operatorname{Exp}([\delta\phi_b]_\times)R_b$. Then
\begin{equation}
 \delta\phi_{b,k+1}=Q_k\!\left(\delta\phi_{b,k}
       +\Delta t\,J_r(\Delta t\,\omega_{b,k})\delta\omega_{b,k}\right),
 \label{eq:state-update-differential}
\end{equation}
where $J_r$ is the right Jacobian of $SO(3)$. The Euler-XYZ maps convert
these spatial attitude variations to coordinate variations, completing
$A_k$ and $C_k$.

The sensitivity of the complete configuration to the decision variables,
$S_k=\partial q_k/\partial z$, follows the recursion
\begin{equation}
 S_0=0,\qquad S_{k+1}=A_kS_k+C_kU_k.
 \label{eq:sensitivity-recursion}
\end{equation}
A single recursive sensitivity rollout supplies all task and clearance
derivatives.

Let $E_{p,k}$ map coordinate variations $\delta q_k$ to end-effector
translation, and let $E_{R,k}\delta q_k=\delta\phi_e$ be the spatial angular
variation defined by $R'_e=\operatorname{Exp}([\delta\phi_e]_\times)R_e$.
Thus the angular Jacobian columns are
$\bigl((\partial R_e/\partial q_j)R_e^T\bigr)^\vee$.
Differentiating the pose errors gives
\begin{equation}
 \frac{\partial e_{p,k}}{\partial z}=E_{p,k}S_k,\quad
 \frac{\partial e_{R,k}}{\partial z}=-J_r^{-1}(e_{R,k})E_{R,k}S_k.
 \label{eq:terminal-pose-jacobian}
\end{equation}
Here $J_r^{-1}$ converts angular variations into logarithmic orientation-error
coordinates. Let $J_{c_i,k}$ be the sphere-center Jacobian and define
$n_{io,k}=(c_i-c_o)/\|c_i-c_o\|_2$ and
$n_{ij,k}=(c_i-c_j)/\|c_i-c_j\|_2$, with centers evaluated at sample $k$.
The collision rows are
\begin{equation}
 \thickmuskip=2mu
 \medmuskip=2mu
 \tfrac{\partial\delta_{io,k}}{\partial z}=n_{io,k}^{T}J_{c_i,k}S_k,\;
 \tfrac{\partial\delta_{ij,k}}{\partial z}=n_{ij,k}^{T}(J_{c_i,k}-J_{c_j,k})S_k.
 \label{eq:collision-jacobian-rows}
\end{equation}
The same sensitivities accumulate the objective gradient. Sparse assembly
uses segment support and the local dependence of acceleration bounds on
adjacent control points.

%% file: texs/09-algorithm.tex
\begin{figure*}[!t]
  \centering
  \begin{subfigure}[t]{0.192\textwidth}
    \includegraphics[width=\linewidth]{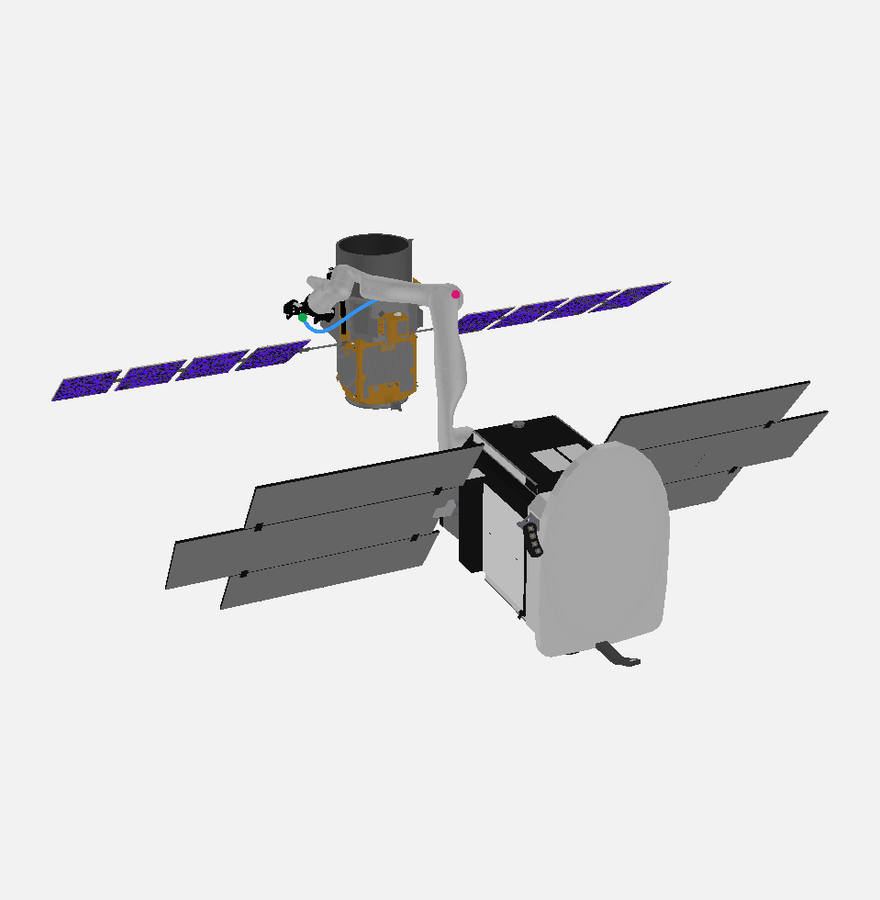}
    \caption{CALIPSO}
  \end{subfigure}\hfill
  \begin{subfigure}[t]{0.192\textwidth}
    \includegraphics[width=\linewidth]{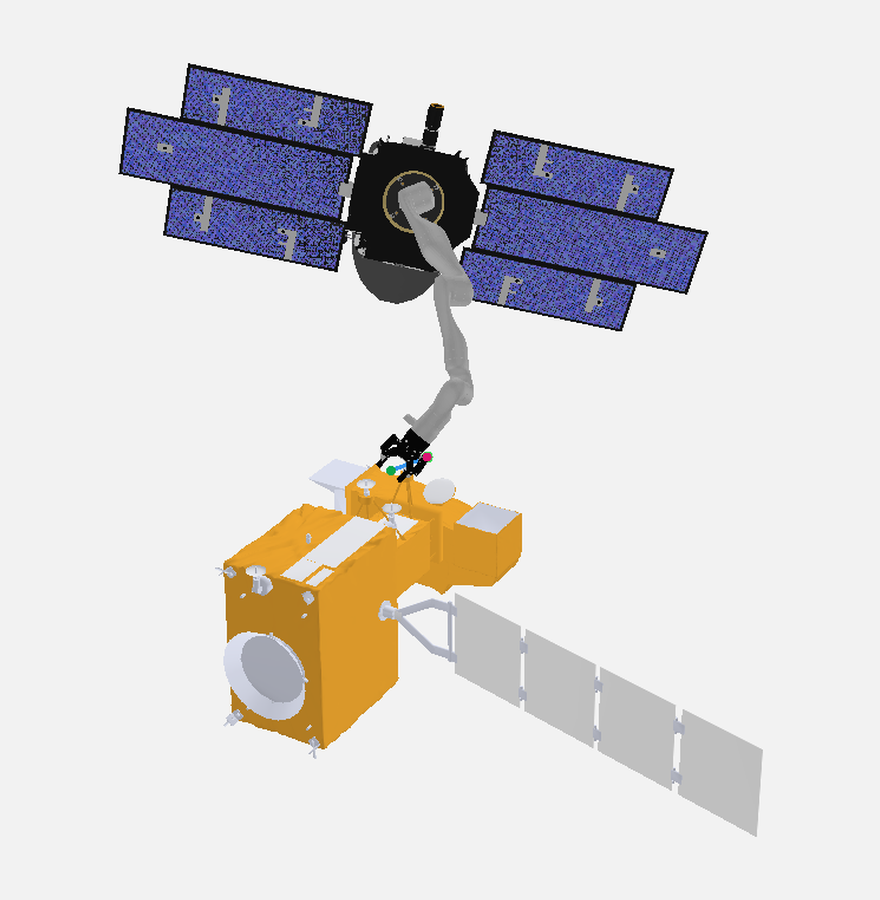}
    \caption{Landsat 7}
  \end{subfigure}\hfill
  \begin{subfigure}[t]{0.192\textwidth}
    \includegraphics[width=\linewidth]{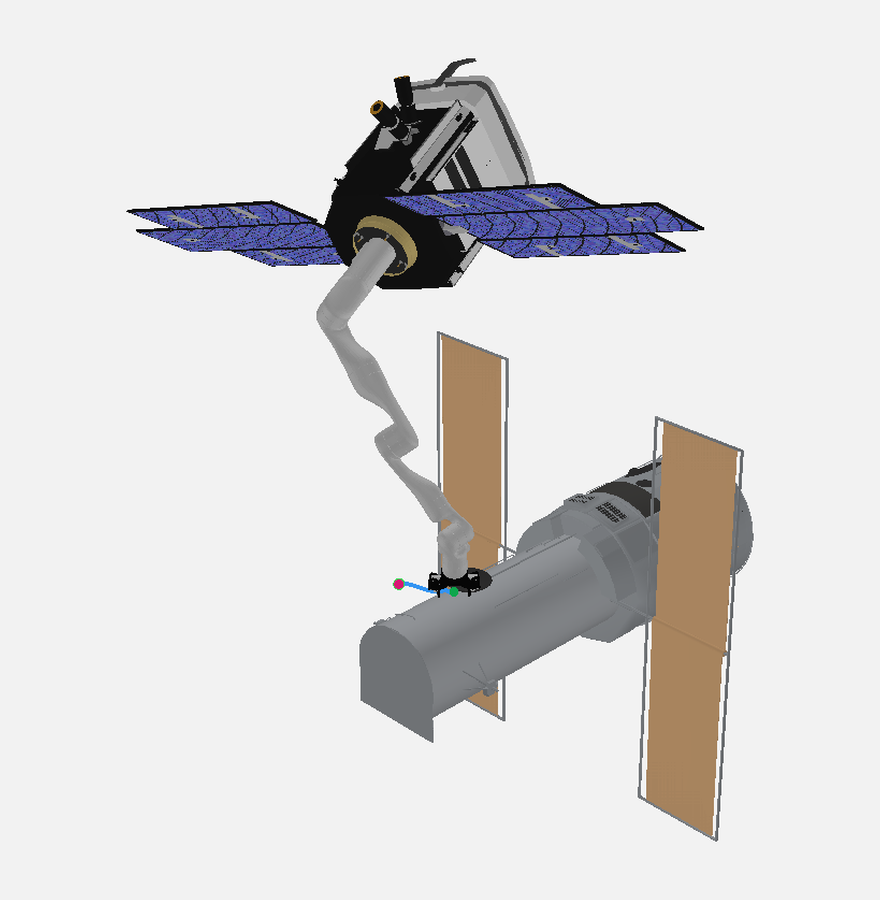}
    \caption{Hubble}
  \end{subfigure}\hfill
  \begin{subfigure}[t]{0.192\textwidth}
    \includegraphics[width=\linewidth]{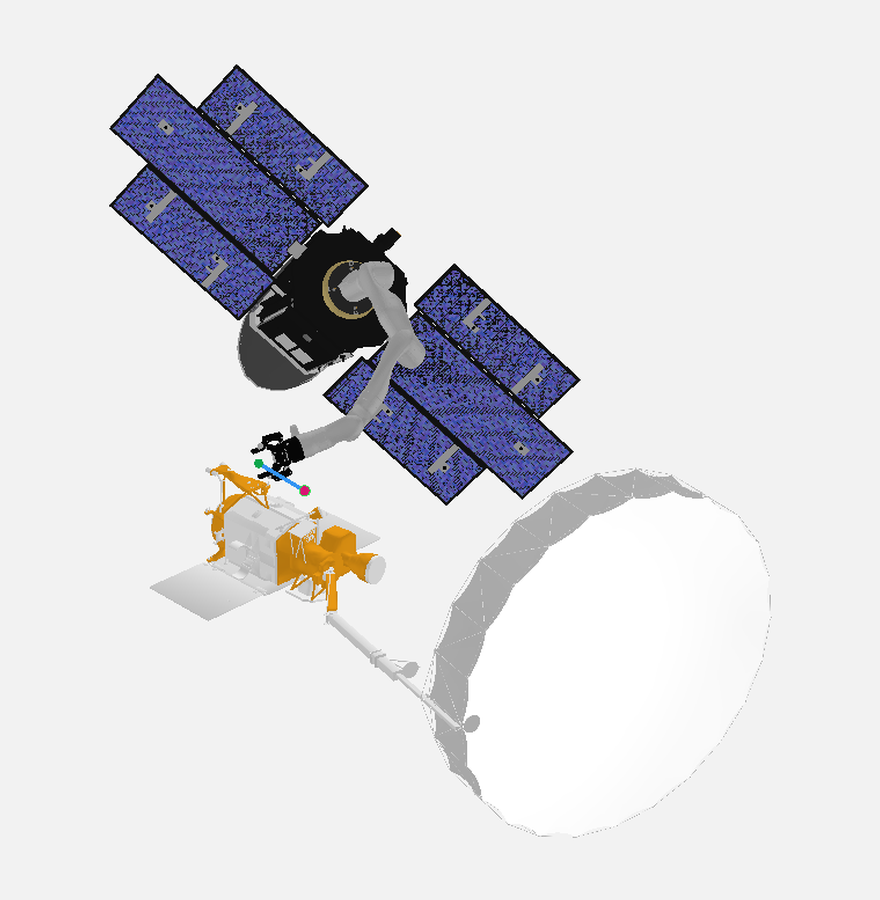}
    \caption{SMAP}
  \end{subfigure}\hfill
  \begin{subfigure}[t]{0.192\textwidth}
    \includegraphics[width=\linewidth]{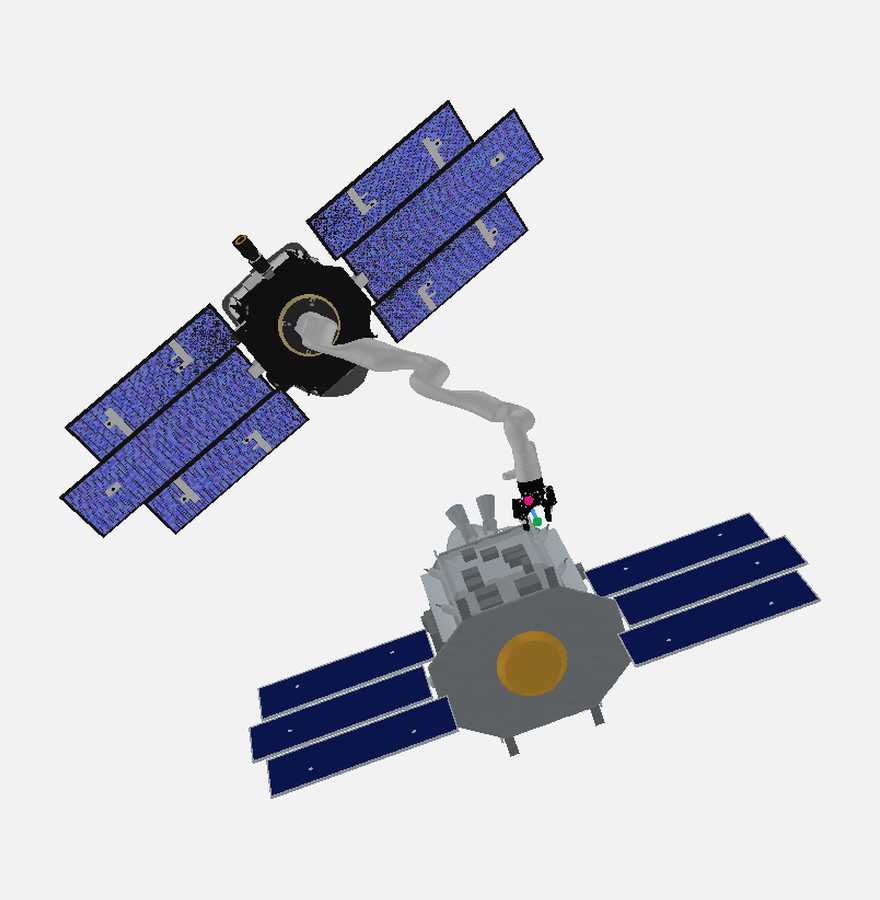}
    \caption{QuikSCAT}
  \end{subfigure}
  \caption{Point-to-point scenes with five target spacecraft and a CloudSat
  chaser. Initial robot configurations are shown with the planned
  end-effector paths and start/arrival markers.}
  \label{fig:spacecraft-scenarios}
\end{figure*}

\subsection{Optimization and Collision Discovery}

Algorithm~\ref{alg:ipopt-bezier-procedure} summarizes FAVOR's stop-on-success mode.
The reduced problem is solved with IPOPT's interior-point filter line-search
method and a limited-memory Hessian approximation
\cite{wachter2006implementation}. The optimizer receives the B\'ezier
parameters and analytical derivatives; the state trajectory remains internal
to the rollout.

Collision discovery uses incremental constraint generation
\cite{hauser2021semiinfinite}. Each update adds the eligible pair--sample
constraint with the smallest signed separation below an activation
distance, then reoptimizes from the current trajectory. Discovery queries
a subset of rollout samples; a final collision check covers every
integration sample. Section~\ref{sec:evaluation} specifies the activation
distance and sampling strides.

\begin{algorithm}[t]
\small
\KwIn{$q_0$, $u_{\mathrm{initial}}$, $(P_0,H_0)$, task poses, arrival time $T$, obstacle motion}
\KwOut{Joint-velocity curve $u(t)$ and its rollout}
$z\leftarrow\operatorname{Initialize}(q_0,u_{\mathrm{initial}},P_0,H_0,\text{task poses},T)$\;
\If{tracking seed passes all feasibility checks}{
  \Return{its velocity curve and rollout}\;
}
$\mathcal A\leftarrow\emptyset$\;
\While{planning budget remains}{
  $\mathcal Q\leftarrow\operatorname{Rollout}(q_0,z)$\;
  $\mathcal C\leftarrow\operatorname{DiscoverClosePairs}(\mathcal Q)$\;
  $\mathcal A\leftarrow\operatorname{UpdateActiveSet}(\mathcal A,\mathcal C)$\;
  $z\leftarrow\operatorname{Optimize}(z,\mathcal A)$\;
  \If{solver, terminal pose, and collision checks over all samples pass}{
    \textbf{break}\;
  }
}
$u(t)\leftarrow\operatorname{Bezier}(z)$\;
$\mathcal Q\leftarrow\operatorname{Rollout}(q_0,z)$\;
\caption{FAVOR: joint-velocity optimization with recursive sensitivities.}
\label{alg:ipopt-bezier-procedure}
\end{algorithm}

\subsubsection{Initialization}
All three tasks use a generalized-Jacobian rollout for initialization.
The end-effector twist is $\xi_e=J_g(q)u+\xi_0(q;P_0,H_0)$, where
$\xi_0$ is the momentum-induced drift \cite{misra2017task}.
Bounded least squares with damping $\lambda$ selects $u$ toward the next guide
pose, using translation differences and rotation logarithms divided by
$\Delta t$; the full state is then propagated. Tracking follows its supplied
reference. Endpoint guides use linear position interpolation and quaternion
SLERP with a quintic time law; interception targets the pre-grasp pose at the
prescribed arrival time. The sampled velocities are
fitted by minimizing $\sum_k\|U_kz-\hat u_k\|^2$ subject to the initial-velocity
boundary and coefficient velocity, acceleration, and continuity constraints. Subsequent endpoint
optimization remains free to change the route.
Tracking can return a feasible fitted seed directly or continue refining
within the budget, retaining the best validated trajectory.

\subsubsection{Parallel Jacobians and Recursive Assembly}
\label{sec:parallel-recursive-assembly}

For a candidate parameter vector $z$, we first sample the joint-velocity
curve and propagate the coupled state sequentially using
\eqref{eq:nonlinear-rollout}. This gives the nominal states $q_0,\ldots,q_N$
and controls $u_0,\ldots,u_{N-1}$, including the accumulated base reaction.
Dependence between successive time samples is handled by this propagation
and the sensitivity recursion described below.

Once the states and controls are available, each local dynamics derivative
$A_k=\partial F/\partial q_k$ and $C_k=\partial F/\partial u_k$ depends only
on its sample's quantities. We therefore evaluate these matrices in parallel
across samples. The end-effector state Jacobians $E_{p,k}$ and $E_{R,k}$
are evaluated in the same way. The collision stage likewise parallelizes
sphere placement, sphere-center state Jacobians, and BVH queries across the
requested samples. These are derivatives with respect to the state and
instantaneous control at a sample.

A forward pass then applies \eqref{eq:sensitivity-recursion} to obtain
$S_k=\partial q_k/\partial z$. The term $A_kS_k$ carries the effect of earlier
commands, while $C_kU_k$ adds the current command's dependence on the control
points. The map $U_k=\partial u_k/\partial z$ follows directly from the
Bernstein basis and continuity map. Each update reuses $S_k$, requiring
$N$ matrix updates instead of $N(N+1)/2$ when every sample's sensitivity is
rebuilt independently from the initial state. Finally,
\eqref{eq:terminal-pose-jacobian} and \eqref{eq:collision-jacobian-rows}
compose the local geometric derivatives with $S_k$ to form the task and
clearance Jacobians with respect to $z$. State-dependent objective terms use
the same sensitivities; control penalties differentiate directly through the
Bernstein representation. The thread-scaling study measures the time for the
local dynamics and pose-Jacobian stage, while the independent-assembly
ablation in Table~\ref{tab:computational-ablation} reconstructs each $S_k$
without reusing preceding sensitivities.

\subsubsection{Shared Trajectory Caching}
\label{sec:trajectory-caching}

IPOPT requests objective values, constraint values, and their derivatives
through separate callbacks, often at the same parameter vector. For a fixed
task and initial state, we share one nominal rollout across these requests.
The first value evaluation stores the complete state sequence and derived
pose quantities, even if that callback requires only the terminal pose.
A subsequent derivative request evaluates local Jacobians at the stored
states and attaches pose Jacobians and propagated sensitivities to the same
trajectory, avoiding another state rollout. Entries are reused only when the
control coefficients and arrival time match exactly.

We retain separate entries for the most recent derivative evaluation and
value-only trial evaluation. Line-search trials can replace the latter
without discarding the trajectory and sensitivities at the current
linearization point. Repeated value and derivative requests at that point
therefore reuse the same computations.

Collision data are cached by trajectory sample, including sphere placements,
robot BVHs, and sphere-center Jacobians when requested. Changing the control
coefficients or arrival time invalidates these collision entries.
The Bernstein basis and prescribed obstacle BVHs are independent of those
coefficients and remain available across IPOPT iterates. The caching
ablation in Table~\ref{tab:computational-ablation} removes reuse across
callbacks while retaining this static precomputation.

%% file: texs/12-evaluation.tex
\subsection{Experimental Setup}

The simulator couples a seven-DoF Kinova Gen3-derived arm and Robotiq
2F-85 gripper to a freely translating and rotating base with CloudSat
geometry. It includes link kinematics, masses, inertias, and self- and
environment-collision models. We evaluate endpoint planning, inspection
tracking, and prescribed-time pre-grasp interception.

For endpoint planning and interception, FAVOR enforces the prescribed initial
joint velocity and arm-position limits of $[-2\pi,2\pi]$~rad. Base
coordinates are unbounded.

\begin{table}[t]
  \caption{Initial base motion.}
  \label{tab:initial-motion}
  \centering\footnotesize
  \setlength{\tabcolsep}{3pt}
  \input{texs/table_initial_motion}
\end{table}

\begin{table}[t]
  \caption{Point-to-point planning. $S$: success rate.}
  \label{tab:point-to-point-results}
  \centering\footnotesize
  \setlength{\tabcolsep}{0.9pt}
  \input{texs/table_point_to_point}
\end{table}

Experiments use an Intel Core i9-14900KF CPU with IPOPT for
trajectory optimization and Gurobi for QP. \mbox{FAVOR}'s analytical Jacobian
and collision computations use 24 CPU threads; the spline baseline uses
24 collision threads. Gurobi uses 24 threads.
Reported computation covers the complete task; $T$ denotes physical duration.
Point-to-point planning and tracking use shared trajectory caching;
interception retains separate value and derivative caches. The Jacobian
initializer uses $\lambda=10^{-3}$. Initialization counts toward the planning budget.

Main benchmarks use quintic segments with $C^0$ velocity continuity.
In \eqref{eq:objective-components}, SI units are used and
$(w_u,w_a)=(10^{-3},10^{-4})$ throughout. Terminal weights
$(w_p^T,w_R^T)$ are $(1000,100)$ for point-to-point planning and tracking,
and $(100,10)$ for interception. Tracking uses
$(w_p^r,w_R^r,w_b)=(1000,100,100)$; these weights are zero for endpoint tasks.

Collision discovery uses a 1~m activation distance to select nearby
sphere pairs for constraint generation. The required clearance is
$m_{\mathrm{col}}=20$~mm. Discovery queries every fifth rollout sample for
point-to-point planning and every second sample for tracking and
interception, always including the terminal sample. Successful plans pass
replay checks for velocity and acceleration bounds, clearance, momentum
balance, and task-specific pose criteria at every integration sample.

Initial base velocities (Table~\ref{tab:initial-motion}) are defined in the
base frame and rotated into the inertial frame for each starting
configuration. V0 is rest.

\begin{figure}[!tp]
  \centering
  \sbox0{\includegraphics{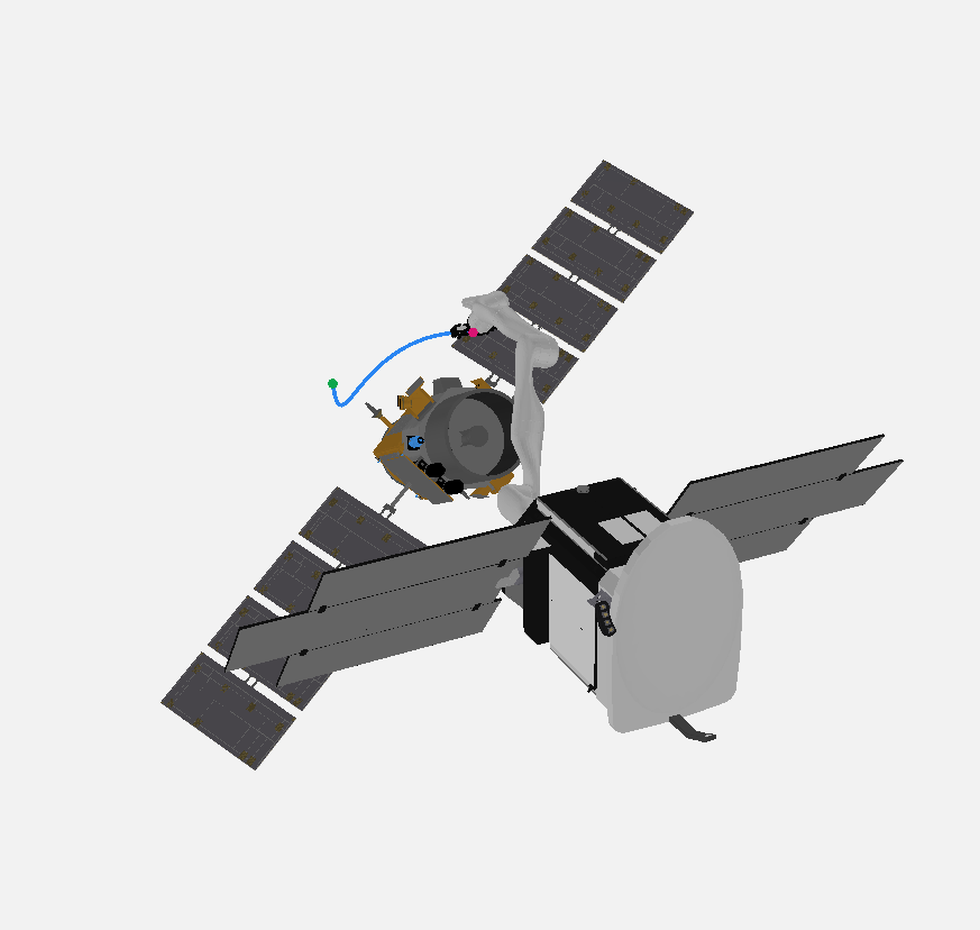}}
  \edef\trackingtoptrim{\the\dimexpr\ht0/7\relax}
  \edef\trackingbottomtrim{\the\dimexpr\ht0/5\relax}
  \begin{subfigure}[t]{0.48\columnwidth}
    \centering
    \includegraphics[width=0.95\linewidth,
      trim=0pt \trackingbottomtrim{} 0pt \trackingtoptrim{},clip]{figures/scenario_tracking_1.png}
    \caption{Moving-target STA approach}
  \end{subfigure}\hfill
  \begin{subfigure}[t]{0.48\columnwidth}
    \centering
    \includegraphics[width=0.95\linewidth,
      trim=0pt \trackingbottomtrim{} 0pt \trackingtoptrim{},clip]{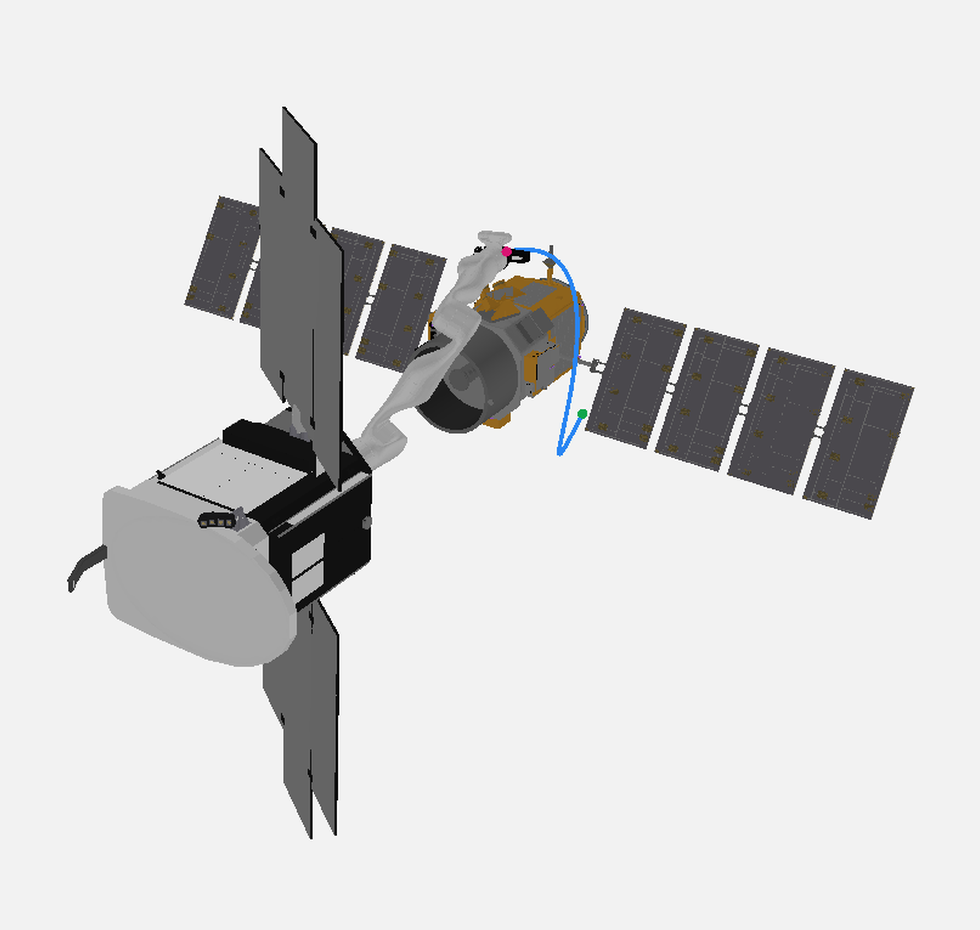}
    \caption{IIR-to-STA transfer}
  \end{subfigure}
  \par\smallskip
  \begin{subfigure}[t]{0.48\columnwidth}
    \centering
    \includegraphics[width=0.95\linewidth,
      trim=0pt \trackingbottomtrim{} 0pt \trackingtoptrim{},clip]{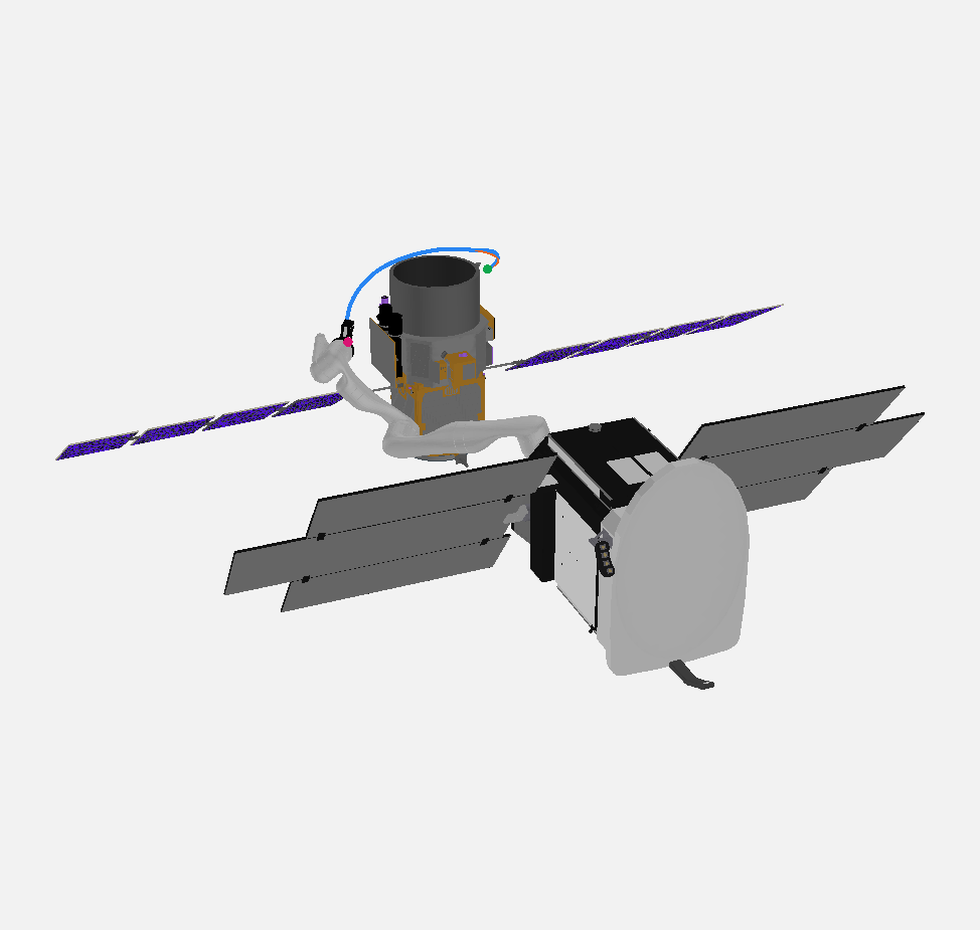}
    \caption{ILR telescope-ring bypass}
  \end{subfigure}\hfill
  \begin{subfigure}[t]{0.48\columnwidth}
    \centering
    \includegraphics[width=0.95\linewidth,
      trim=0pt \trackingbottomtrim{} 0pt \trackingtoptrim{},clip]{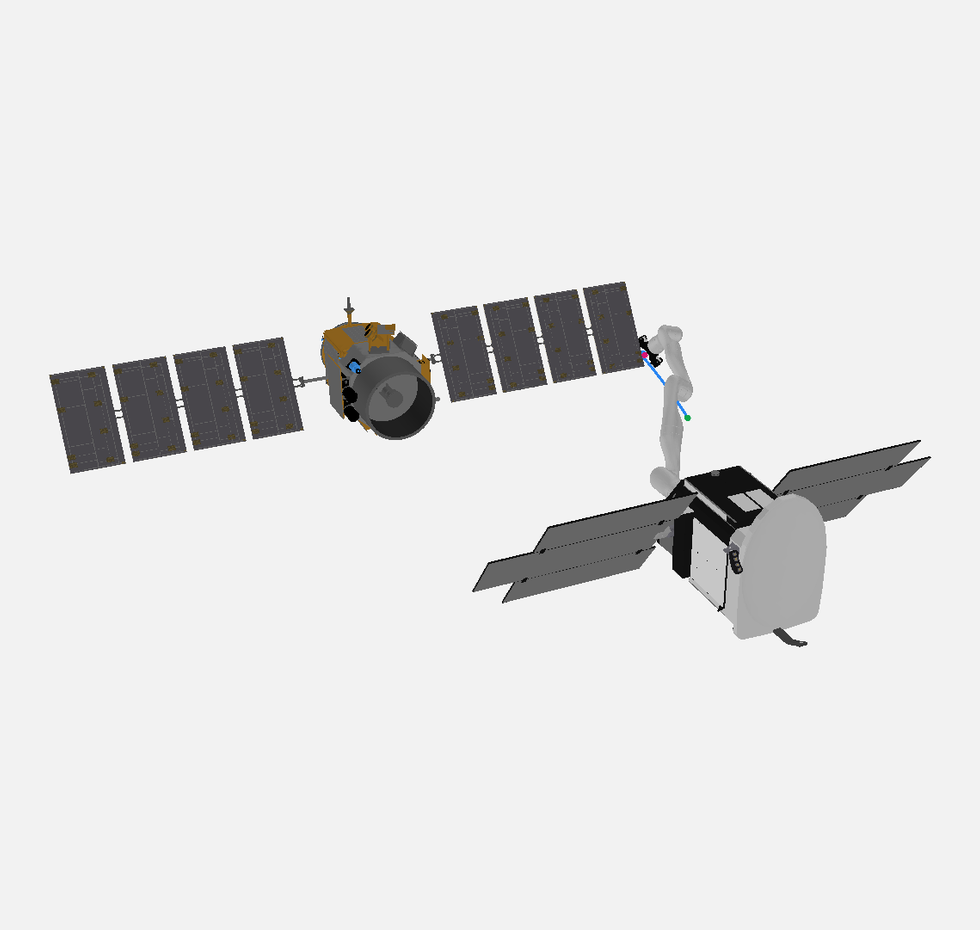}
    \caption{Lateral IIR approach}
  \end{subfigure}
  \caption{CALIPSO inspection tasks at zero initial base motion and mass ratio
  ten. Blue and orange curves show the planned and reference paths;
  markers identify start and arrival. The target moves in (a) and is
  static in (b)--(d).}
  \label{fig:tracking-scenarios}
\end{figure}

\subsection{Point-to-Point Motion Planning}

Five geometric tasks around each of CALIPSO, Landsat~7, Hubble, SMAP,
and QuikSCAT are crossed with initial-motion profiles V0--V4 and
base-to-arm mass ratios $\{20,14,10,5,2\}$, giving \PointTotal{} cases
(125 per spacecraft). Scaling base mass and inertia together varies the
base reaction while retaining each task's geometry. The target bodies and
appendages in Fig.~\ref{fig:spacecraft-scenarios} use FOAM sphere unions
with 512, 204, 256, 165, and 248 spheres, respectively. All methods use a
\PointIpoptBudget~s planning budget; mean computation includes all attempts,
including failures and timeouts.

FAVOR uses one quintic segment. CALIPSO maneuvers have
$T=5.10$--$6.52$~s and $\Delta t=0.01$~s; the other spacecraft use $T=6$~s and
$\Delta t=0.02$~s. Joint-velocity and acceleration bounds are
$2\pi/3$~rad/s and 5~rad/s$^2$, with terminal pose tolerances of 0.005~m and
0.05~rad. The in-house
Jacobian-steered kinodynamic RRT receives at most 20,000 expansions
and one recorded seed per instance. It propagates the coupled state, projects
commands onto the control bounds, and checks eight fractions of each edge.

The adapted position-spline baseline \cite{lampariello2013grasping} uses
IPOPT and clamped uniform quartic B-splines with seven control points per
joint. Fixing the initial position and velocity leaves 35 free variables,
matching FAVOR.
Spline differentiation supplies joint velocities for the same dynamics,
objective, bounds, collision geometry, and success criteria. The baseline uses finite-difference nonlinear
derivatives and fixed body-pair clearance constraints every fifth integration
step and at the terminal state. The random variant seeds the spline toward
a joint configuration sampled within the joint limits
\cite{lampariello2013grasping}; the terminal-IK variant uses an IK estimate
for the desired end-effector pose. Each performs one IPOPT
trajectory-optimization run per instance, with initialization charged to
the budget. FAVOR uses warm-started collision discovery with up to
1000 updates within the shared budget.
Discovered constraints remain active throughout each point-to-point solve.

FAVOR achieves \PointIpoptSuccess/\PointTotal{} successes
(\PointIpoptRate\%) with \PointIpoptTime~s mean computation
(Table~\ref{tab:point-to-point-results}), compared with \PointSplineRate\%
and \PointSplineTime~s for the IK-initialized position-spline baseline.
One CALIPSO maneuver misses the
terminal position tolerance; all tasks on the other four spacecraft succeed.
Successful plans have mean terminal errors of \PointPositionMeanMM~mm and
\PointOrientationMean~rad.

FAVOR's IK variant initializes the velocity coefficients from the joint
displacement to a terminal IK solution divided by the maneuver duration,
clipped to the velocity limits. With the same planner settings and
\PointIpoptBudget~s budget, it yields \PointIkSuccess/\PointTotal{} successes
and \PointIkTime~s mean time. On \PointInitializationShared{} shared successes,
Jacobian initialization reduces mean time from \PointIkSharedTime~s to
\PointJacobianSharedTime~s.

The Jacobian-steered RRT completes none of the cases within the budget.
Its steering requires pose distance to decrease at each step, while acceleration bounds
limit command changes to 0.05 or 0.10~rad/s. Projection and collision
rejection restrict progress.

\subsection{Reference Trajectory Tracking}

\begin{table}[!t]
  \caption{Trajectory tracking.}
  \label{tab:tracking-results}
  \centering\footnotesize
  \setlength{\tabcolsep}{3pt}
  \input{texs/table_tracking}
\end{table}

\begin{figure}[!t]
  \centering
  \begin{subfigure}[t]{0.48\columnwidth}
    \begin{overpic}[width=\linewidth]{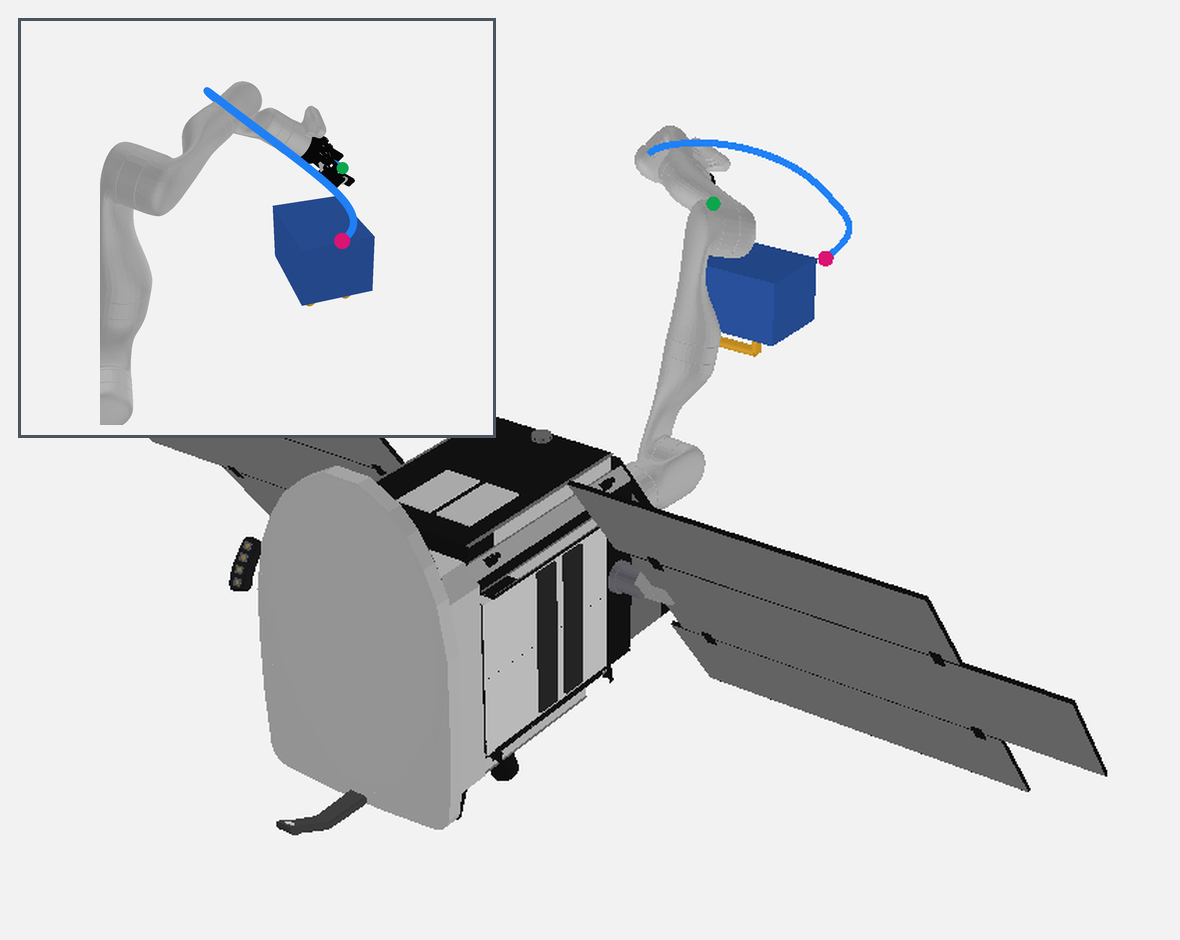}
      \put(98,76){\makebox(0,0)[tr]{\colorbox{white}{\footnotesize $t=0~\mathrm{s}$}}}
    \end{overpic}
  \end{subfigure}\hfill
  \begin{subfigure}[t]{0.48\columnwidth}
    \begin{overpic}[width=\linewidth]{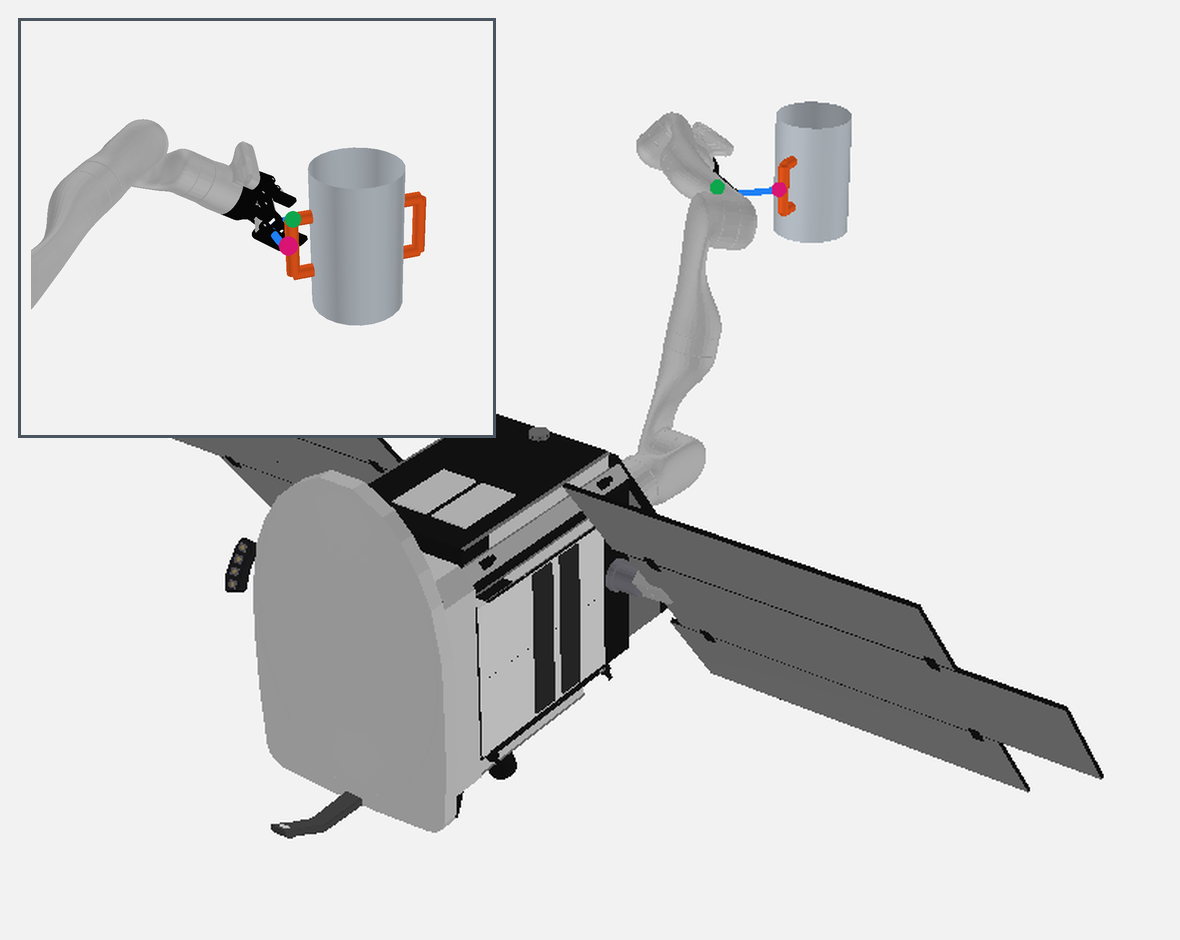}
      \put(98,76){\makebox(0,0)[tr]{\colorbox{white}{\footnotesize $t=0~\mathrm{s}$}}}
    \end{overpic}
  \end{subfigure}
  \par\smallskip
  \begin{subfigure}[t]{0.48\columnwidth}
    \begin{overpic}[width=\linewidth]{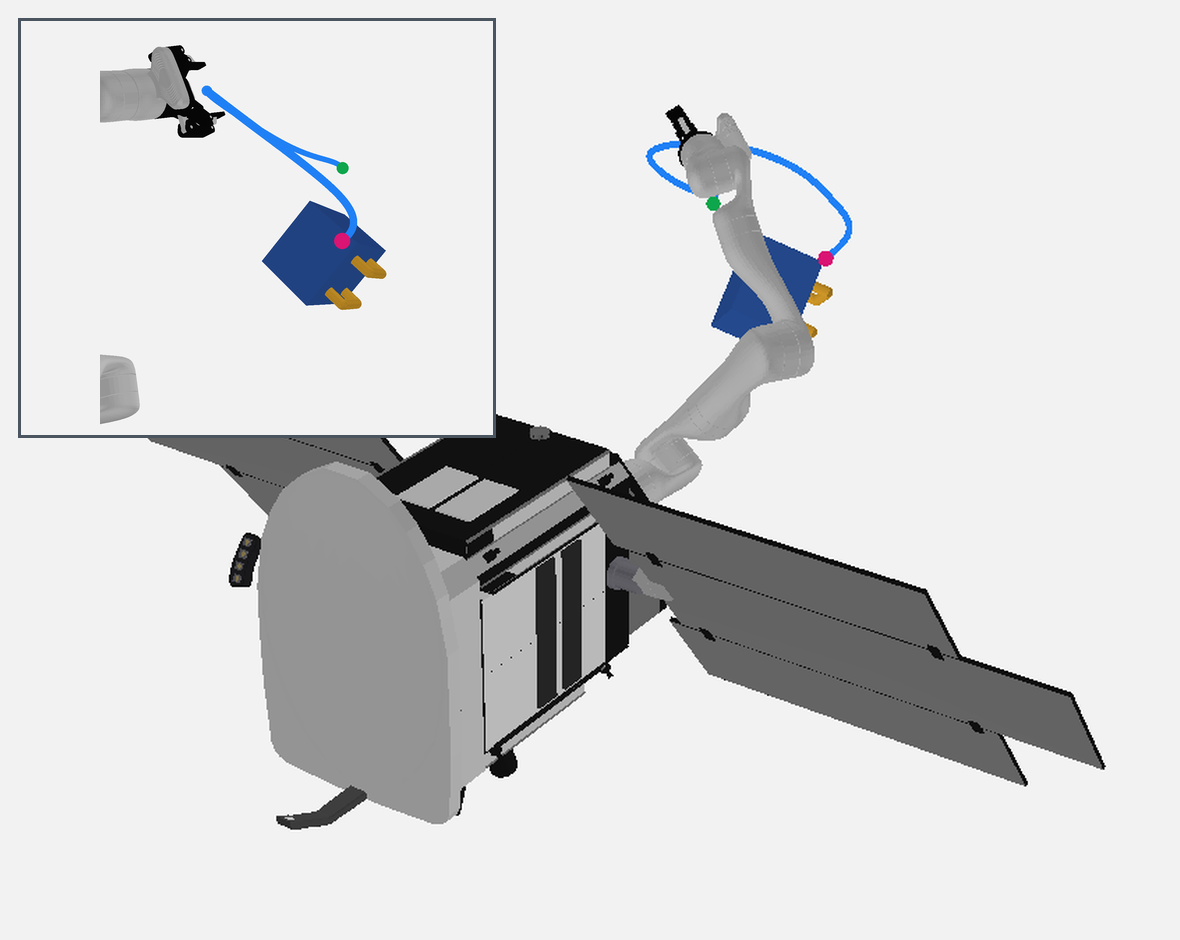}
      \put(98,76){\makebox(0,0)[tr]{\colorbox{white}{\footnotesize $t=5~\mathrm{s}$}}}
    \end{overpic}
  \end{subfigure}\hfill
  \begin{subfigure}[t]{0.48\columnwidth}
    \begin{overpic}[width=\linewidth]{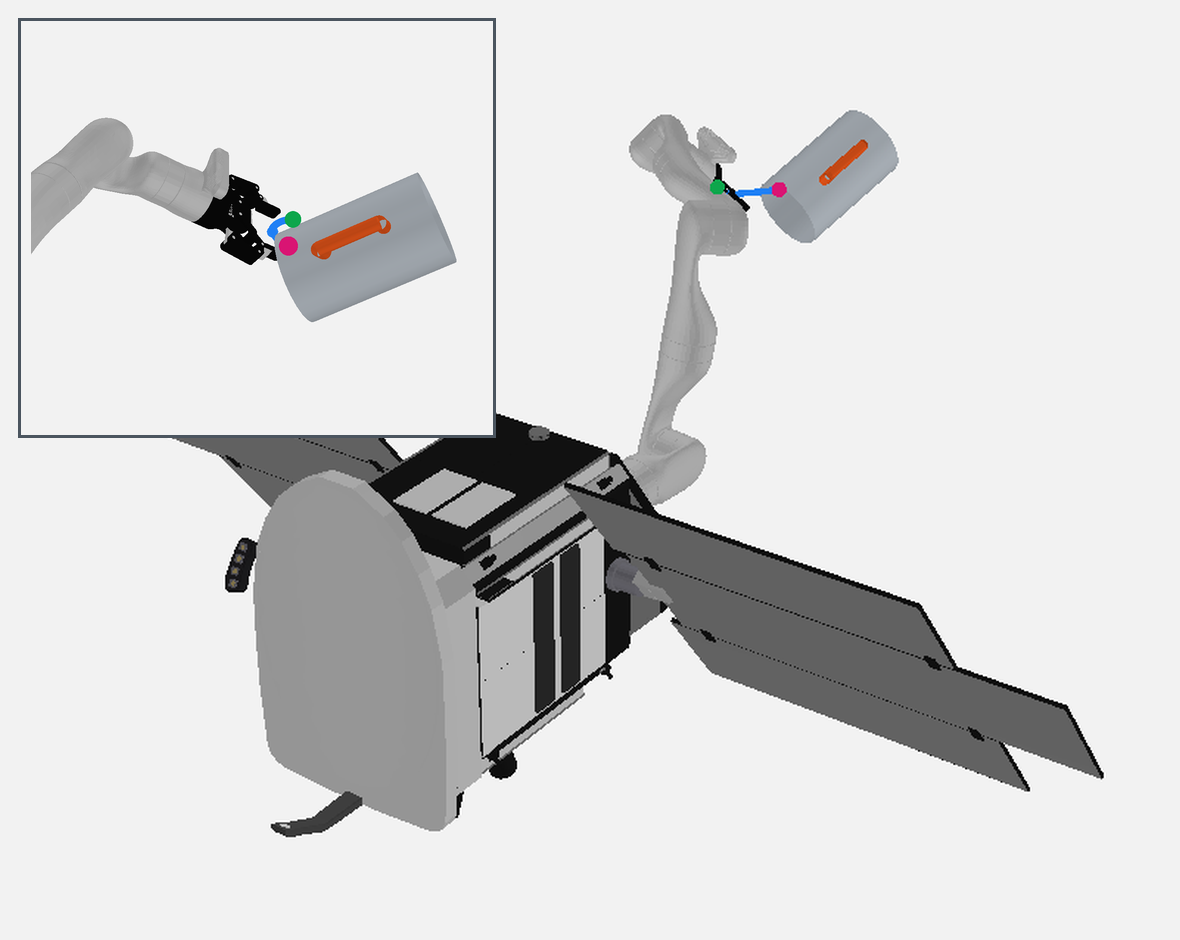}
      \put(98,76){\makebox(0,0)[tr]{\colorbox{white}{\footnotesize $t=5~\mathrm{s}$}}}
    \end{overpic}
  \end{subfigure}
  \par\smallskip
  \begin{subfigure}[t]{0.48\columnwidth}
    \begin{overpic}[width=\linewidth]{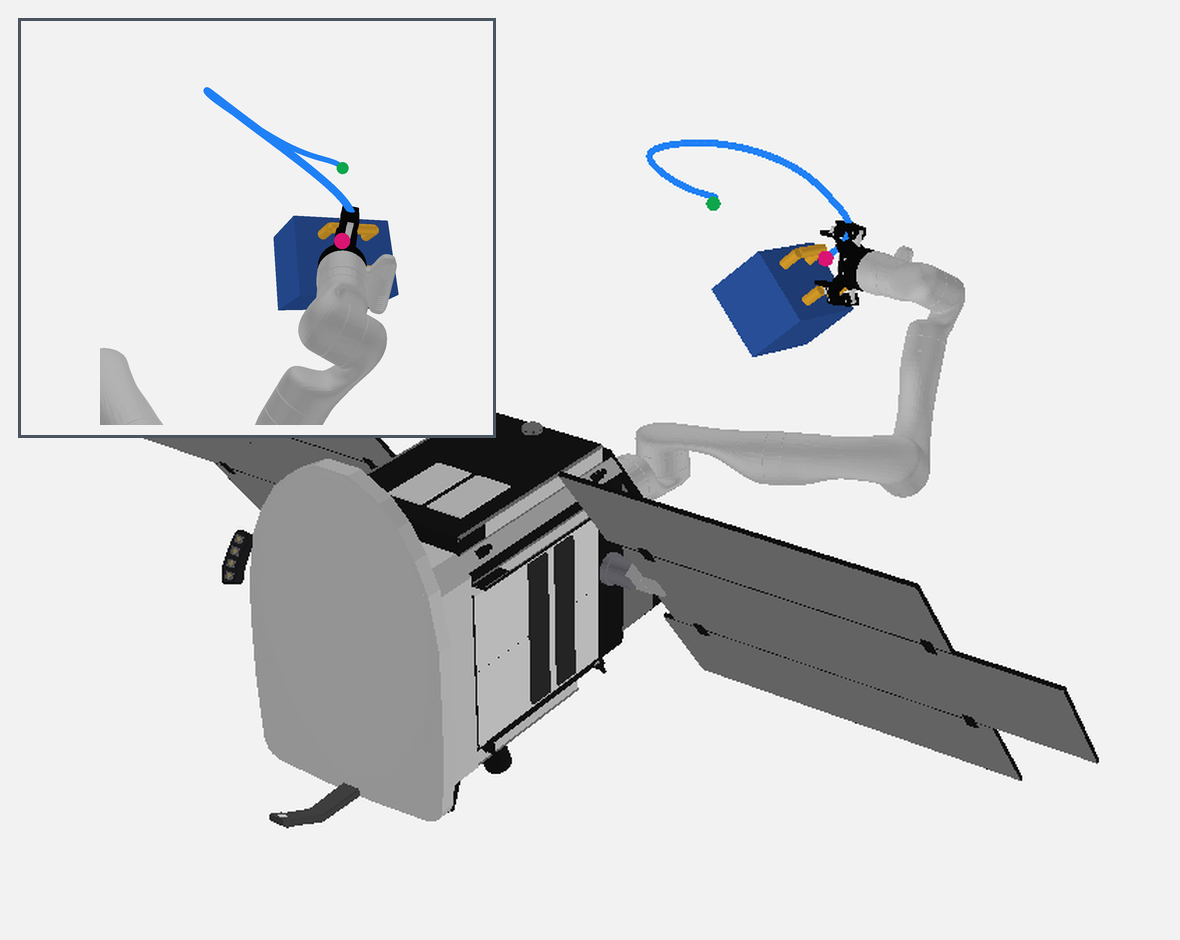}
      \put(98,76){\makebox(0,0)[tr]{\colorbox{white}{\footnotesize $t=10~\mathrm{s}$}}}
    \end{overpic}
  \end{subfigure}\hfill
  \begin{subfigure}[t]{0.48\columnwidth}
    \begin{overpic}[width=\linewidth]{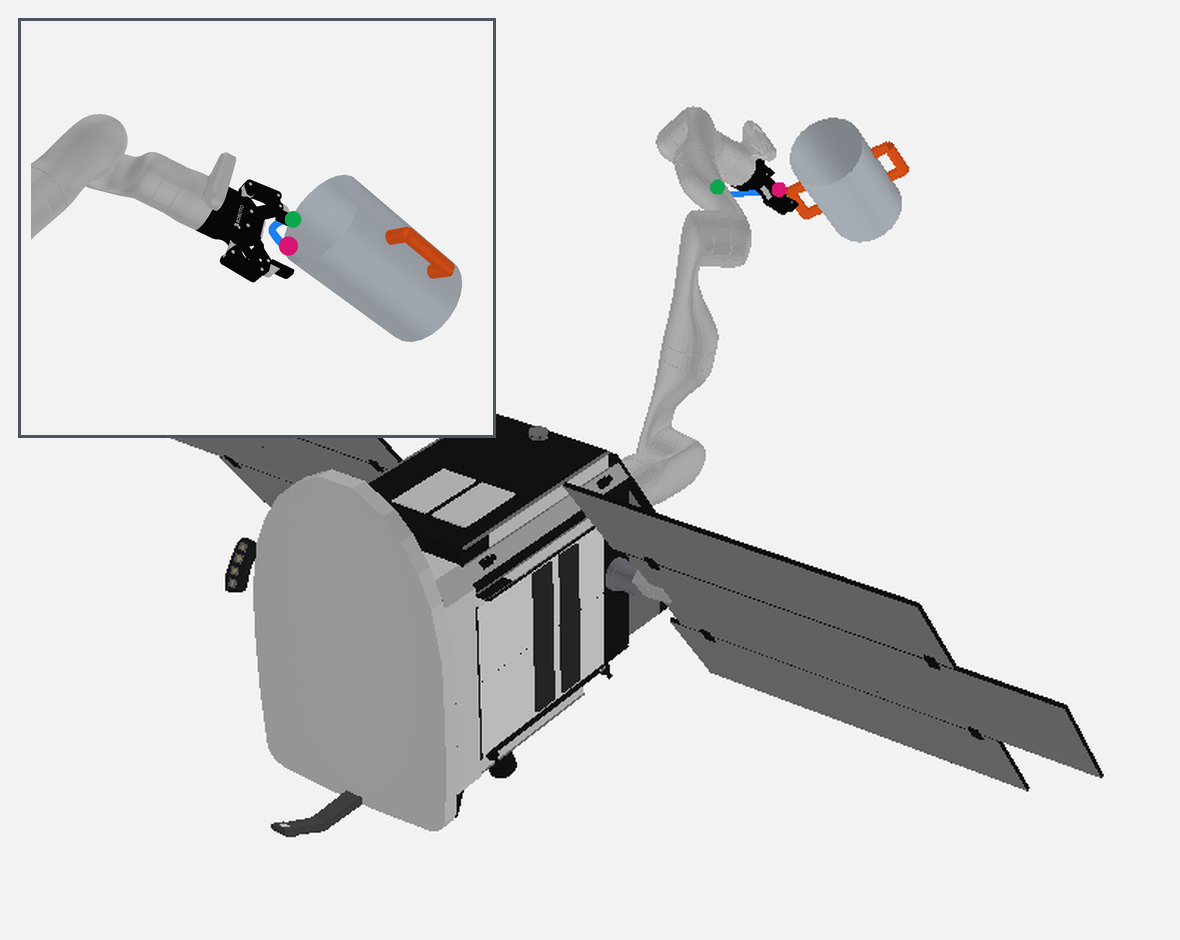}
      \put(98,76){\makebox(0,0)[tr]{\colorbox{white}{\footnotesize $t=10~\mathrm{s}$}}}
    \end{overpic}
  \end{subfigure}
  \caption{Timed pre-grasp approaches under compound tumble. The left and
  right columns show the payload box and canister, respectively. Corner
  insets provide a closer view of the end effector and object.
  Blue curves show the planned paths; green and magenta mark start and arrival.}
  \label{fig:interception-scenarios}
\end{figure}

\begin{figure}[!t]
  \centering
  \begin{subfigure}[t]{0.24\columnwidth}
    \includegraphics[width=\linewidth]{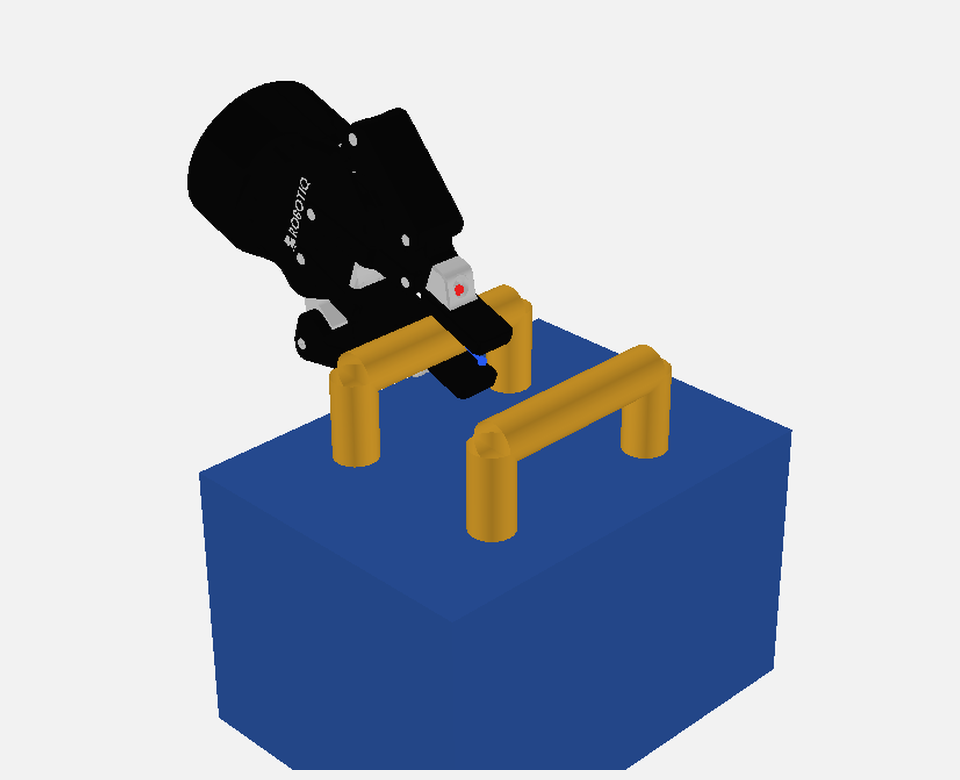}
  \end{subfigure}\hfill
  \begin{subfigure}[t]{0.24\columnwidth}
    \includegraphics[width=\linewidth]{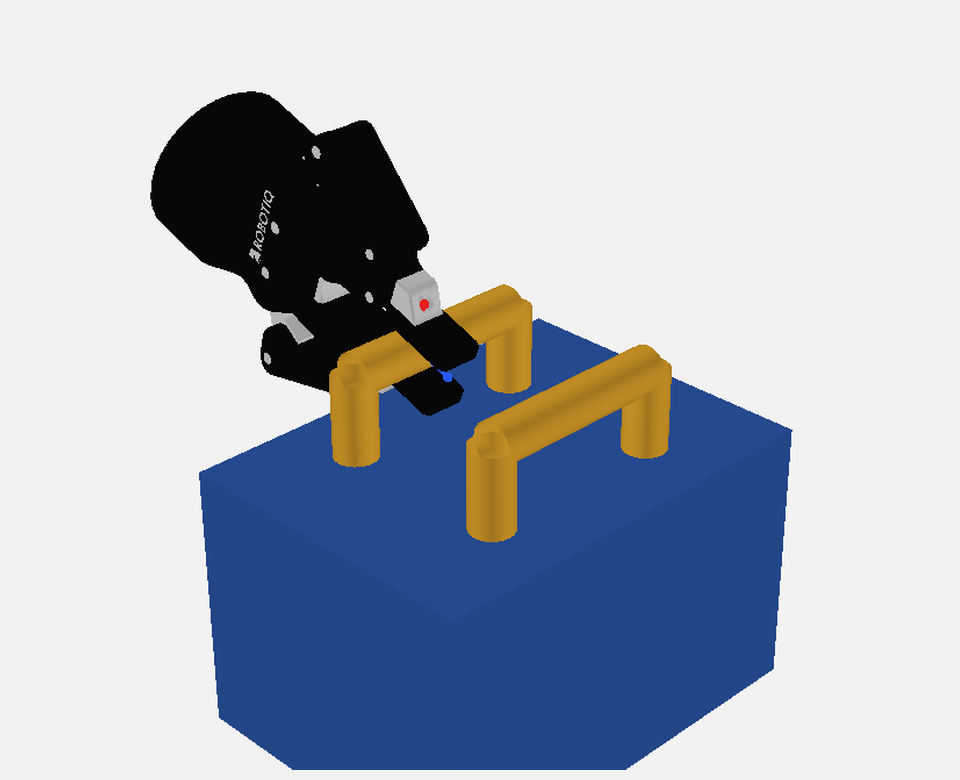}
  \end{subfigure}\hfill
  \begin{subfigure}[t]{0.24\columnwidth}
    \includegraphics[width=\linewidth]{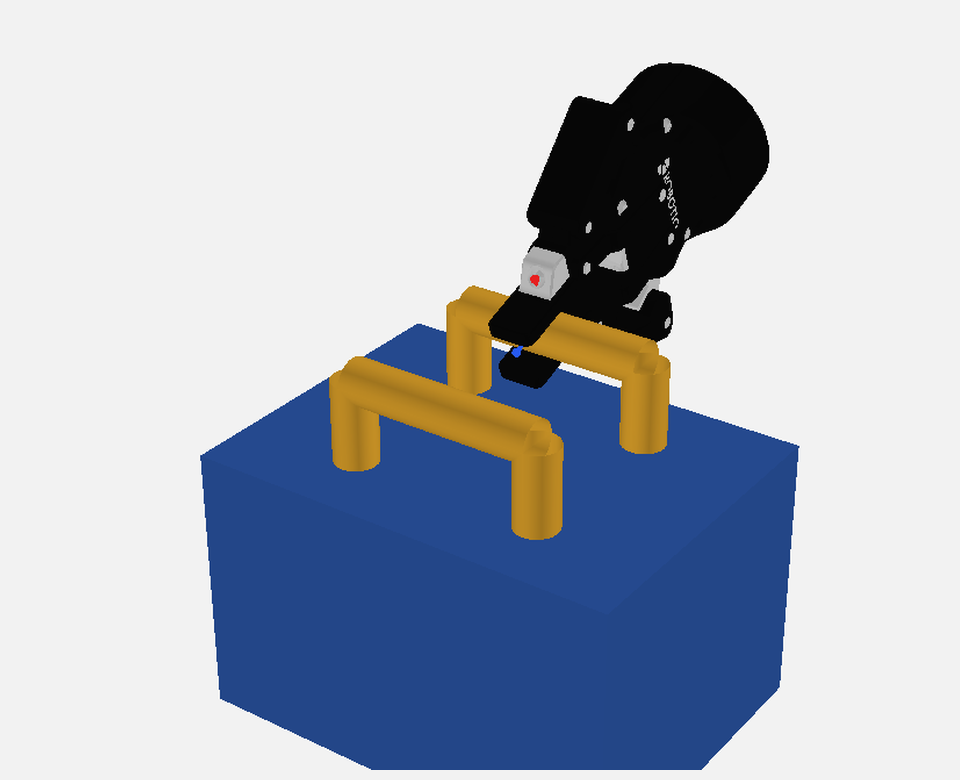}
  \end{subfigure}\hfill
  \begin{subfigure}[t]{0.24\columnwidth}
    \includegraphics[width=\linewidth]{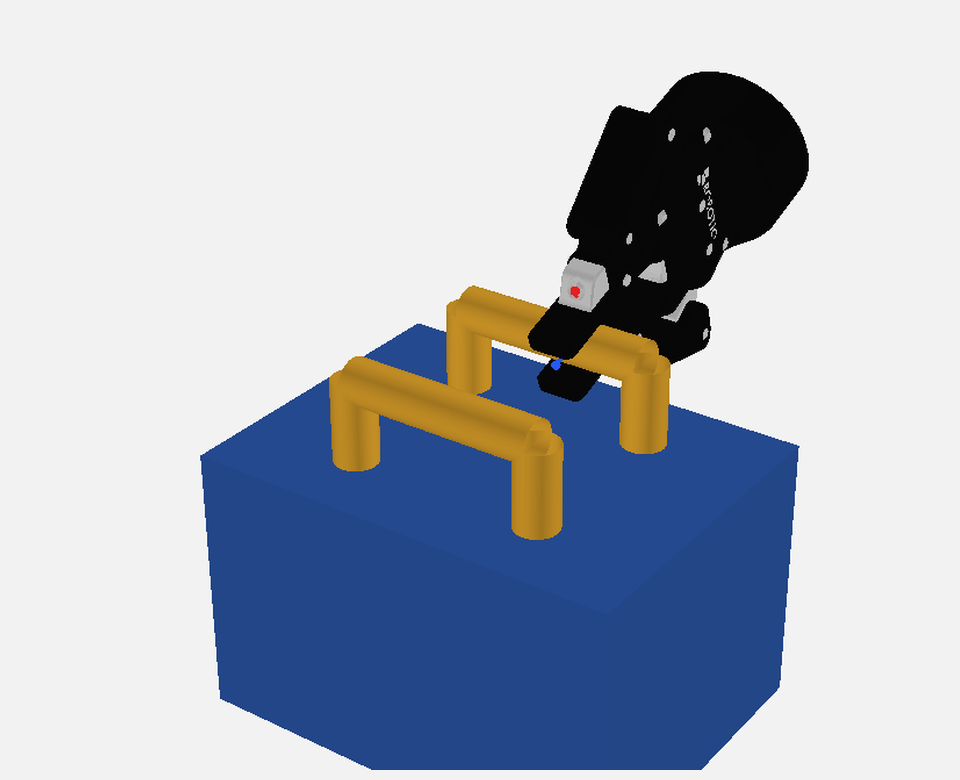}
  \end{subfigure}
  \caption{Illustrative Robotiq 2F-85 contact grasps.}
  \label{fig:grasp-candidates}
\end{figure}

\begin{figure}[!tp]
  \centering
  \includegraphics[width=\columnwidth]{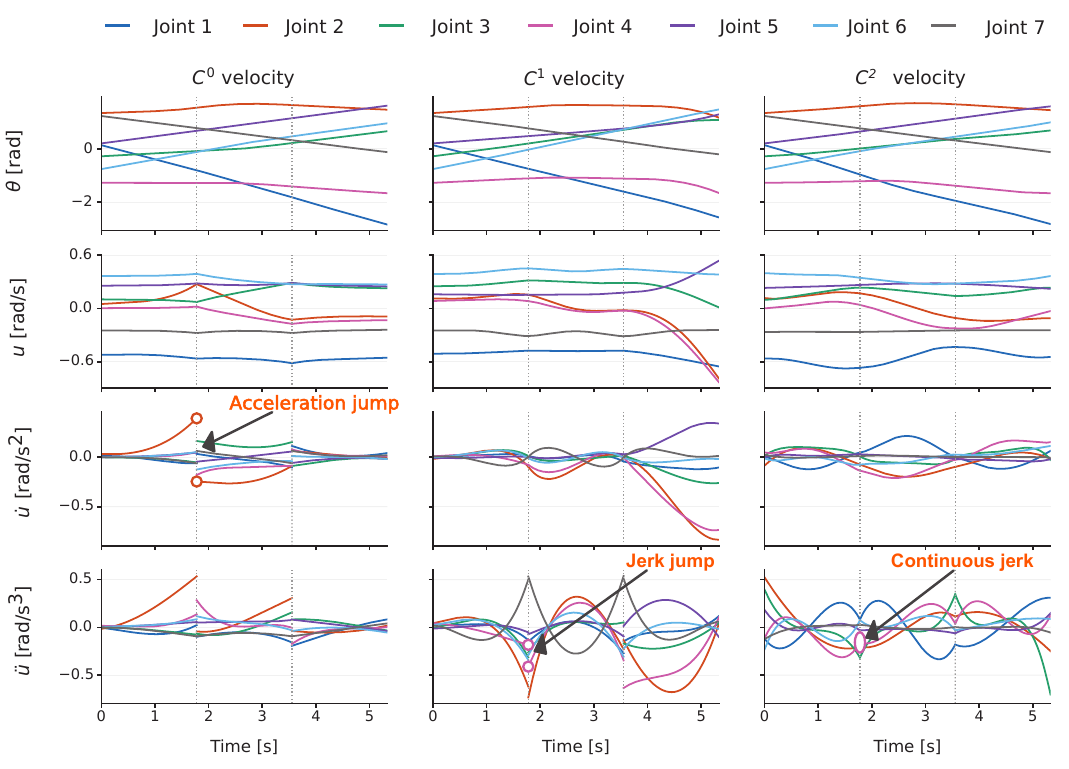}
  \caption{The same CALIPSO goal with three quintic segments and $C^0$, $C^1$,
  or $C^2$ velocity continuity. Rows show seven-joint position, velocity,
  acceleration, and jerk. Dotted lines mark joins; callouts highlight
  derivative continuity. Jerk is evaluated within segments.}
  \label{fig:trajectory-profiles}
\end{figure}

The four CALIPSO references in Fig.~\ref{fig:tracking-scenarios} come
from the Space Manipulator Inspection Benchmarks \cite{xbai2026benchmark}.
They target CALIPSO's STA, IIR, and ILR regions.
Three have static targets and one has prescribed target motion. Crossing
each with profiles V0--V2 and mass ratios $\{2,10,20\}$ gives 27 static and
nine dynamic cases.

We compare with the original single-step generalized-Jacobian QP
\cite{misra2017task}. Both methods use the same propagated state model over
$T=50$~s with a 0.1~s step. Joint-velocity and acceleration bounds are
$\pi/15$~rad/s and $\pi/12$~rad/s$^2$. The evaluated single-step QP does not impose
acceleration continuity.
FAVOR uses four quintic
segments (140 free parameters) and a \TrackingBudget~s search budget. Stop returns
the first feasible plan; refine uses the remaining budget. We also evaluate a terminal-IK variant with refinement under the same
budget, using the initial configuration as a fallback when IK fails. Tracking success
requires terminal
position and orientation errors within 0.05~m and 0.15~rad. The inspection
corridor additionally bounds their 95th percentiles by 0.10~m and 0.25~rad,
and their maxima by 0.20~m and 0.50~rad.

Tracking errors are the mean position- and orientation-error norms over
each trajectory.
Table~\ref{tab:tracking-results} averages these errors and whole-trajectory
runtimes equally over the \TrackingStaticShared{} static and
\TrackingDynamicShared{} dynamic cases completed by all four variants;
runtimes include final checks.
Both Jacobian-initialized FAVOR variants complete \TrackingStaticIpoptSuccess/27
static and \TrackingDynamicIpoptSuccess/9 dynamic cases, versus
\TrackingStaticLegacyIkSuccess/27 and \TrackingDynamicLegacyIkSuccess/9 for
the IK variant, and \TrackingStaticQpSuccess/27 and
\TrackingDynamicQpSuccess/9 for QP. Of the 30 stop-on-success completions,
\TrackingSeedSuccess{} are feasible directly after initialization.
Stop-on-success has the lowest computation time. Refinement further
reduces position error, while QP has the lowest static orientation error.
Both Jacobian-initialized variants have lower errors than the IK
variant and lower dynamic errors than QP.

\subsection{Prescribed-Time Pre-Grasp Interception}

\begin{table}[t]
  \vspace{-4mm}
  \caption{Pre-grasp interception. $S$: success rate.}
  \label{tab:interception-results}
  \centering\footnotesize
  \setlength{\tabcolsep}{3pt}
  \input{texs/table_interception}
  \vspace{-3mm}
\end{table}

For interception, FAVOR evaluates the object-relative pre-grasp goal at the
requested arrival time and accounts for the object's motion during the
approach (Fig.~\ref{fig:interception-scenarios}). Two handled objects, three
chaser layouts, three torque-free tumble regimes, and arrival times of 5 and
10~s produce 36 instances. Slow and fast regimes rotate initially about the
highest-inertia principal axis at 0.10 and 0.25~rad/s. Compound tumble uses
initial components $(0.10,0.14,0.18)$~rad/s along principal axes ordered by
increasing inertia.

GraspIt! generates contact poses offline \cite{miller2004graspit}. The open
Robotiq gripper is retreated to a collision-free pre-grasp pose, and fixed-base
inverse kinematics checks candidate reachability.
Fig.~\ref{fig:grasp-candidates} gives additional contact-grasp examples
generated with the Robotiq 2F-85 model. Both planners receive the same four
pre-grasp targets per instance
and a 10~s budget per candidate.
FAVOR uses two quintic segments and the Jacobian
initialization described above.
The rollout step is 0.1~s; velocity and acceleration bounds are 1~rad/s and
5~rad/s$^2$, with terminal pose tolerances of 0.01~m and 0.05~rad.

The position-spline baseline uses twelve control points per joint. Fixing
the initial position and velocity leaves 70 free variables, matching FAVOR.
Each initialization variant performs one IPOPT solve per candidate and
includes initialization in its budget. All methods share the objective,
acceptance criteria, and supplied object-motion history. Collision checks
occur every second integration step during optimization and every step
during validation.

FAVOR solves all \InterceptionTotal{} instances, with
\InterceptionCandidateSuccess{} of 144 candidate attempts succeeding
(Table~\ref{tab:interception-results}). Mean computation per instance is
\InterceptionTimeMean~s, including all four candidate attempts and the
single candidate timeout. The IK-initialized position-spline baseline
completes \InterceptionSplineSuccess{} instances with \InterceptionSplineTime~s
mean computation. Selected trajectories have mean terminal errors
of \InterceptionPositionMeanMM~mm and \InterceptionOrientationMean~rad.

\subsection{Trajectory Profiles and Momentum}

\begin{figure}[t]
  \centering
  \includegraphics[width=\columnwidth]{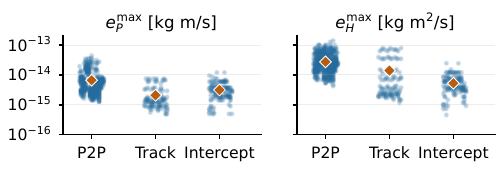}
  \caption{Momentum conservation across all \MomentumTrajectories{} main-benchmark
  trajectories from FAVOR. Diamonds show group means.}
  \label{fig:benchmark-momentum}
\end{figure}

Fig.~\ref{fig:trajectory-profiles} shows prescribed derivative continuity
at segment joins.
Fig.~\ref{fig:benchmark-momentum} reports peak linear and angular momentum
deviations, $e_P^{\max}=\max_k\|P_k-P_0\|_2$ and
$e_H^{\max}=\max_k\|H_k-H_0\|_2$, respectively. Dots represent individual
trajectories: 625 point-to-point runs (P2P), 108 tracking runs (Track),
and 144 interception candidates (Intercept), including failed attempts.
The largest linear and angular deviations
are \MomentumPMaximum~kg\,m/s and \MomentumHMaximum~kg\,m$^2$/s.

\subsection{Computational Performance}

Table~\ref{tab:computational-ablation} compares four FAVOR variants on
\CompCases{} cases spanning 25 geometries, initial-motion profiles, and
mass ratios. All use the same velocity representation, Jacobian
initialization, bounds, collision discovery, and 60~s budget.
\emph{Finite differences} substitutes numerical objective gradients and
nonlinear constraint Jacobians. \emph{No cache} removes cross-callback
reuse; \emph{Independent assembly} rebuilds each sample's sensitivity
from the initial state. The latter two preserve the full method's
trajectories and optimizer histories. Planning means include failures;
assembly means are per full-horizon analytical sensitivity evaluation,
with cases weighted equally.

\begin{table}[!htb]
  \caption{Computational ablation.}
  \label{tab:computational-ablation}
  \centering\footnotesize
  \setlength{\tabcolsep}{3pt}
  \input{texs/table_computational}
\end{table}

Finite differences increase mean planning time
\CompFiniteDifferenceSlowdown{}-fold and reduce successes from
\CompFullSuccess{} to \CompFiniteDifferenceSuccess{}.
In a separate 25-geometry study with fixed IK initialization and 24 collision
threads, using 24 Jacobian threads instead of one reduces full-horizon
local Jacobian evaluation time \CompJacobianSpeedup{}-fold.

%% file: texs/table_initial_motion.tex
\begin{tabular*}{\columnwidth}{@{\extracolsep{\fill}}lcc@{}}
\toprule
 & $v_{b,0}^{b}$ [mm/s] & $\omega_{b,0}^{b}$ [mrad/s] \\
\midrule
V0 & $(0,\,0,\,0)$ & $(0,\,0,\,0)$ \\
V1 & $(-5.073,\,-3.478,\,3.535)$ & $(-10.112,\,14.558,\,-7.211)$ \\
V2 & $(-0.946,\,-0.813,\,0.545)$ & $(-0.984,\,-1.133,\,-1.033)$ \\
V3 & $(2.032,\,-2.484,\,4.063)$ & $(-10.278,\,-6.229,\,-2.592)$ \\
V4 & $(0.434,\,3.983,\,2.342)$ & $(-1.420,\,1.323,\,-1.550)$ \\
\bottomrule
\end{tabular*}

%% file: texs/table_point_to_point.tex
\begin{tabular*}{\columnwidth}{@{\extracolsep{\fill}}lrrrrrrrr@{}}
\toprule
 & \multicolumn{2}{c}{FAVOR} & \multicolumn{2}{c}{Spline (random)} & \multicolumn{2}{c}{Spline (IK)} & \multicolumn{2}{c@{}}{RRT} \\
Spacecraft & $S$ [\%] & Time [s] & $S$ [\%] & Time [s] & $S$ [\%] & Time [s] & $S$ [\%] & Time [s] \\
\midrule
CALIPSO & \textbf{99.2} & \textbf{7.282} & 10.4 & 58.704 & 12.0 & 58.097 & 0.0 & $\geq 60$ \\
Landsat 7 & \textbf{100.0} & \textbf{0.548} & 48.0 & 45.922 & 56.0 & 52.448 & 0.0 & $\geq 60$ \\
Hubble & \textbf{100.0} & \textbf{0.320} & 50.4 & 50.459 & 90.4 & 47.281 & 0.0 & $\geq 60$ \\
SMAP & \textbf{100.0} & \textbf{0.262} & 64.0 & 46.631 & 74.4 & 48.115 & 0.0 & $\geq 60$ \\
QuikSCAT & \textbf{100.0} & \textbf{0.482} & 60.0 & 39.210 & 81.6 & 45.408 & 0.0 & $\geq 60$ \\
\midrule
Overall & \textbf{99.8} & \textbf{1.779} & 46.6 & 48.185 & 62.9 & 50.270 & 0.0 & $\geq 60$ \\
\bottomrule
\end{tabular*}

%% file: texs/table_tracking.tex
\begin{tabular*}{\columnwidth}{@{\extracolsep{\fill}}lrrrr@{}}
\toprule
Method & Success [\%] & $\bar e_p$ [mm] & $\bar e_R$ [rad] & Avg. time [s] \\
\midrule
\multicolumn{5}{@{}l}{\textit{Static}} \\
FAVOR (stop) & \textbf{85.2} & 5.74 & 0.0031 & \textbf{0.293} \\
FAVOR (refine) & \textbf{85.2} & \textbf{3.14} & 0.0103 & 3.968 \\
FAVOR (IK) & \textbf{85.2} & 7.3 & 0.0308 & 4.037 \\
Single-step QP & 37.0 & 4.71 & \textbf{0.00275} & 4.325 \\
\midrule
\multicolumn{5}{@{}l}{\textit{Dynamic}} \\
FAVOR (stop) & \textbf{77.8} & 0.016 & $1.25\!\times\!10^{-5}$ & \textbf{0.304} \\
FAVOR (refine) & \textbf{77.8} & \textbf{0.00713} & $\boldsymbol{8.25\!\times\!10^{-6}}$ & 4.055 \\
FAVOR (IK) & 66.7 & 3.56 & 0.0285 & 4.027 \\
Single-step QP & 55.6 & 0.136 & $5.64\!\times\!10^{-5}$ & 5.767 \\
\bottomrule
\end{tabular*}

%% file: texs/table_interception.tex
\begin{tabular*}{\columnwidth}{@{\extracolsep{\fill}}lrrrrrr@{}}
\toprule
 & \multicolumn{2}{c}{FAVOR} & \multicolumn{2}{c}{Spline (random)} & \multicolumn{2}{c@{}}{Spline (IK)} \\
Object & $S$ [\%] & Time [s] & $S$ [\%] & Time [s] & $S$ [\%] & Time [s] \\
\midrule
Payload box & \textbf{100.0} & \textbf{2.754} & 22.2 & 39.736 & 83.3 & 32.669 \\
Canister & \textbf{100.0} & \textbf{3.332} & 22.2 & 39.617 & 38.9 & 36.978 \\
\midrule
Overall & \textbf{100.0} & \textbf{3.043} & 22.2 & 39.676 & 61.1 & 34.824 \\
\bottomrule
\end{tabular*}

%% file: texs/table_computational.tex
\begin{tabular*}{\columnwidth}{@{\extracolsep{\fill}}lrrr@{}}
\toprule
Variant & Success [\%] & Plan [s] & Assembly [ms] \\
\midrule
FAVOR (full) & \textbf{100.0} & \textbf{1.635} & \textbf{1.924} \\
No cache & \textbf{100.0} & 3.301 & 2.079 \\
Independent assembly & \textbf{100.0} & 2.557 & 8.246 \\
Finite differences & 92.0 & 7.613 & -- \\
\bottomrule
\end{tabular*}

%% file: texs/15-conclusion.tex
We present FAVOR, which combines compact B\'ezier joint-velocity optimization
with analytical recursive sensitivities for prescribed-time free-floating
manipulation. Linear coefficient constraints enforce control bounds and
continuity, while the rollout captures base--arm coupling. Simulated reaching,
tracking, and pre-grasp interception demonstrate higher success than the
evaluated baselines; controlled ablations quantify the computational benefits
of analytical derivatives, caching, and recursive assembly.
The current evaluation assumes known dynamics and prescribed target motion,
with collision checks at integration samples. Future work includes hardware
validation and extending FAVOR to steer kinodynamic RRT~\cite{graesdal2026semidefinite}.